\documentclass[10pt,twocolumn,letterpaper]{article}

\usepackage[pagenumbers]{wacv} %
\usepackage{graphicx}
\usepackage{amsmath}
\usepackage{amssymb}
\usepackage{appendix}
\usepackage{subcaption}
\usepackage{pifont}%
\newcommand{\cmark}{\ding{51}}%
\newcommand{\xmark}{\ding{55}}%

\usepackage{siunitx}
\usepackage[normalem]{ulem}
\robustify\uline
\def\Uline#1{#1\llap{\uline{\phantom{#1}}}} %
\usepackage{booktabs}
\usepackage{multirow}
\usepackage{tabularx, array}
\usepackage{xcolor}
\usepackage{rotating} %
\usepackage[pagebackref,breaklinks,colorlinks]{hyperref}

\usepackage{listings}
\usepackage[capitalize]{cleveref}
\crefname{section}{Sec.}{Secs.}
\Crefname{section}{Section}{Sections}
\Crefname{table}{Table}{Tables}
\crefname{table}{Tab.}{Tabs.}

\def\wacvPaperID{*****} %
\def\confName{WACV}
\def\confYear{2024}

\usepackage[acronym,toc,xindy]{glossaries}
\usepackage{xparse}
\DeclareDocumentCommand{\newdualentry}{ O{} O{} m m m m } {
	\newglossaryentry{gls-#3}{name={#5},text={#5\glsadd{#3}},
		description={#6},#1
	}
	\makeglossaries
	\newacronym[see={[Glossary:]{gls-#3}},#2]{#3}{#4}{#5\glsadd{gls-#3}}
}
\setglossarystyle{list} %
\newacronym{ai}{AI}{artificial intelligence}
\newacronym{xai}{XAI}{explainable \gls{ai}}
\newacronym{ad}{AD}{anomaly detection}
\newacronym{as}{AS}{anomaly segmentation}
\newacronym{dl}{DL}{deep learning}
\newacronym{ml}{ML}{machine learning}
\newacronym{ood}{OOD}{out-of-distribution}
\newacronym{qc}{QC}{quality control}
\newacronym{kd}{KD}{knowledge distillation}
\newacronym{rl}{RL}{reinforcement learning}
\newacronym{gan}{GAN}{generative adversarial network}
\newacronym{cgan}{cGAN}{conditional \gls{gan}}
\newacronym{ae}{AE}{autoencoder}
\newacronym{vae}{VAE}{variational autoencoder}
\newacronym{aae}{AAE}{adversarial autoencoder}
\newacronym{gmm}{GMM}{Gaussian mixture model}
\newacronym{ccd}{CCD}{charge-coupled device}
\newacronym{roi}{ROI}{region of interest}
\newacronym{svm}{SVM}{support vector machine}
\newacronym{smo}{SMO}{sequential minimal optimization}
\newacronym{fpn}{FPN}{feature pyramid networks}

\newacronym{roc}{ROC}{receiver operating characteristic}
\newacronym{pr}{PR}{precision-recall}
\newacronym{auroc}{AU\glsentryshort{roc}}{area under the \gls{roc} curve}
\newacronym{aupr}{AU\glsentryshort{pr}}{area under the \gls{pr} curve}
\newacronym{auprosavg}{AU\glsentryshort{prosavg}}{area under the \gls{prosavg} curve} %
\newacronym{aupros}{AU\glsentryshort{pros}}{area under the \gls{pros} curve}
\newacronym{prosavg}{\glsentryshort{pr}\textsubscript{os,avg}}{class-averaged, \gls{pros}}
\newacronym{pros}{\glsentryshort{pr}\textsubscript{os}}{open set \gls{pr}}
\newacronym{iou}{IoU}{intersection over union}
\newacronym{tp}{TP}{true positive}
\newacronym{tn}{TN}{true negative}
\newacronym{fp}{FP}{false positive}
\newacronym{fn}{FN}{false negative}
\newacronym{tpr}{TPR}{true positive rate}
\newacronym{tnr}{TNR}{true negative rate}
\newacronym{fnr}{FNR}{false negative rate}
\newacronym{fpr}{FPR}{false positive rate}
\newacronym{pro}{PRO}{per-region-overlap}
\newacronym{flop}{FLOP}{floating-point operation}
\newacronym{fps}{FPS}{frames per second} %
\newacronym{cca}{CCA}{connected component analysis} %
\newacronym{lpips}{LPIPS}{learned perceptual image patch similarity}
\newacronym{nst}{NST}{neural style transfer}

\newacronym{pdf}{PDF}{probability density function}
\newacronym{cdf}{CDF}{cumulative distribution function}
\newacronym{gde}{GDE}{Gaussian density estimation}

\newacronym{mse}{MSE}{mean squared error}

\newacronym{cnn}{CNN}{convolutional neural network}
\newacronym{pca}{PCA}{principal component analysis}  %
\newacronym{fc}{FC}{fully connected}
\newacronym{relu}{ReLU}{rectified linear unit}
\newacronym{serlu}{SERLU}{scaled exponentially-regularized linear unit}
\newacronym{oas}{OAS}{oracle approximating shrinkage}  %
\newacronym{npca}{NPCA}{negated \gls{pca}}
\newacronym{sed}{SED}{standardized Euclidean distance}
\newacronym{ann}{ANN}{artificial neural network}
\newacronym{mlp}{MLP}{multi-layer perceptron}
\newacronym{nlp}{NLP}{natural language processing}

\newacronym{gda}{GDA}{Gaussian discriminant analysis}
\newacronym{nn}{NN}{nearest neighbor}
\newacronym{1nn}{1-\glsentryshort{nn}}{1-\glsentrylong{nn}}
\newacronym{ocsvm}{oc-SVM}{one-class \gls{svm}}
\newacronym{svdd}{SVDD}{support vector data description}
\newacronym{if}{IF}{isolation forest}
\newacronym{lof}{LOF}{local outlier factor}
\newacronym{odin}{ODIN}{outlier detection using in-degree number}
\newacronym{osr}{OSR}{open set recognition}
\newacronym[longplural={known known classes}]{kkc}{KKC}{known known class}
\newacronym[longplural={known unknown classes}]{kuc}{KUC}{known unknown class}
\newacronym[longplural={unknown unknown classes}]{uuc}{UUC}{unknown unknown class}
\newacronym[longplural={unknown known classes}]{ukc}{UKC}{unknown known class}
\newacronym{msp}{MSP}{maximum softmax probability}
\newacronym{oe}{OE}{outlier exposure}
\newacronym{vrm}{VRM}{vicinal risk minimization}
\newacronym{ebs}{EBS}{energy-based score}
\newacronym{nll}{NLL}{negative log-likelihood}
\newacronym{em}{EM}{expectation maximization}
\newacronym{bic}{BIC}{Bayesian information criterion}
\newacronym{sem}{SCEM}{scanning electron microscopy}
\newacronym{maxlogit}{MaxL}{maximum of the unnormalized predictions} %
\newacronym{fssd}{FSSD}{feature space singularity distance}
\newacronym{maha}{Maha}{Mahalanobis}
\newacronym{db}{d\&b}{dyed \& bleached}
\newacronym{avi}{AVI}{automated visual inspection}
\newacronym{avis}{AVIS}{\gls{avi} system}
\newacronym{aif}{AiF}{German  Federation  of Industrial  Research  Associations}
\newacronym{en}{EN}{EfficientNet}
\newacronym{ci}{CI}{confidence interval}
\newacronym{ewc}{EWC}{elastic weight consolidation}
\newacronym{nf}{NF}{normalizing flows}
\newacronym{vit}{ViT}{vision transformer}
\newacronym{dpi}{DPI}{dots per inch}
\newacronym{bmbf}{BMBF}{German Federal Ministry of Education and Research}
\newacronym{cagr}{CAGR}{compound annual growth rate}
\newacronym{oscr}{OSCR}{open set class-wise recall}
\newacronym{pet}{PET}{polyethylene terephthalate}
\newacronym{olp}{OLP}{OnLoomPattern}
\newacronym{mida}{MIDA}{multiple image defect analysis}
\newacronym{lda}{LDA}{linear discriminant analysis}
\newacronym{shap}{SHAP}{shapley additive explanation}
\newacronym{csd}{CSD}{cosine similarity distance}
\newacronym{pac}{PAC}{probably approximately correct}
\newacronym{ce}{CE}{cross-entropy}
\newacronym{map}{mAP}{mean average-precision}
\newacronym{spade}{SPADE}{semantic pyramid anomaly detection}
\newacronym{pq}{PQ}{panoptic quality}
\newacronym{sq}{SQ}{segmentation quality}
\newacronym{rq}{RQ}{recognition quality}
\newacronym{fcdd}{FCDD}{fully convolutional data descriptor}
\newacronym{padim}{PaDiM}{patch distribution modeling}

\newdualentry{tsne}{t-SNE}{t-distributed stochastic neighbor embedding}{
	A non-linear visualization technique for high dimensional data \cite{Maaten2008Visualizingdatausing}.
	The algorithm optimizes the Kullback-Leibler divergence between the probabilities of picking nearby points in the low-dimensional visualization with a probability distribution modeled via \gls{gde} in the high-dimensional data
}

\newdualentry{clahe}{CLAHE}{contrast limited adaptive histogram equalization}{
	A processing technique to increase image contrast.
	Histogram equalization is performed on a local neighborhood of each pixel, transforming them with their local \gls{cdf}.
	The histogram is furthermore clipped before computing the \gls{cdf} to limit the amplification of noise in near-constant regions
}

\newdualentry{loo}{LOO}{leave-one-out}{
	A method proposed to increase the robustness of insights generated by training \gls{ml} models on small-scale datasets $\mathcal{D}$.
	Specifically, a single sample of $\mathcal{D}$ is left out for testing, the \gls{ml} model training on the remainder of $\mathcal{D}$, and then tested on the left out sample.
	This procedure is repeated for every single sample contained in $\mathcal{D}$, and the performance measures aggregated
}

\newdualentry{ssim}{SSIM}{structural similarity}{
	A similarity measure between two grayscale images that more closely resembles human perception compared to the \gls{mse}.
	The measure is calculated on sliding, 11~x~11~px windows in the images $\vec{x}$ and $\vec{y}$ as
		\begin{equation*}
		\operatorname{SSIM}(\vec{x}, \vec{y}) = \frac{\left(2\mu_x\mu_y + C_1\right)\left(2\sigma_{xy} + C_2\right)}{\left(\mu_x^2 + \mu_y^2 + C_1\right)\left(\sigma_x^2 + \sigma_y^2 + C_2\right)}.
		\end{equation*}
	Here, $\mu_x, \mu_y$ are the local averages of $\vec{x}$ and $\vec{y}$, $\sigma_x^2, \sigma_y^2$ their local variances, $\sigma_{xy}$ is the local covariance of $\vec{x}$ and $\vec{y}$, and $C_1, C_2$ are constants \cite{Wang2004Imagequalityassessment}
}

\newdualentry{sgd}{SGD}{stochastic gradient descent}{
	In \gls{sgd}, gradient descent is performed on gradients which are computed via backpropagation on mini-batches of the complete dataset \cite{Goodfellow2016Deeplearning}.
	As such, it is a very popular optimizer used in training \gls{dl} models
}

\newdualentry{bce}{BCE}{binary cross-entropy}{
	The \gls{bce} is defined as \begin{equation*}
		\operatorname{\gls{bce}}\left(\Phi(\vec{x}; \mathcal{W}), y\right) = - y\log(\Phi(\vec{x}; \mathcal{W})) - (1 - y) \log(1 - \Phi(\vec{x}; \mathcal{W})),
	\end{equation*}
	where $\Phi$ denotes a neural network parametrized by $\mathcal{W}$ applied to an image $\vec{x}$, and $y$ denotes whether an image is considered normal ($y = 0$) or anomalous ($y = 1$) \cite{Hastie2009ElementsStatisticalLearning}
}

\newdualentry{hsc}{HSC}{hypersphere classifier}{
	The \gls{hsc} is defined as \begin{equation*}
		\begin{split}
			\operatorname{\gls{hsc}}\left(\Phi(\vec{x}; \mathcal{W}), y\right) = &(1 - y)\norm{\Phi(\vec{x}; \mathcal{W})}^2 \\
			& - y \log\left(1 - \exp(-\norm{\Phi(\vec{x}; \mathcal{W})}^2\right),
		\end{split}
	\end{equation*}
	and was proposed to enforce the clustering of normal data in latent representations for \gls{ad} \cite{Ruff2020RethinkingAssumptionsDeep}.
	The anomaly score of an image $\vec{x}$ is then $\norm{\Phi(\vec{x}; \mathcal{W})}^2$
}

\newdualentry{fl}{FL}{focal loss}{
	The \gls{fl} is defined as \begin{equation*}
		\begin{split}
			\operatorname{\gls{fl}}\left(\Phi(\vec{x}; \mathcal{W}), y\right) = &- y (1 - \Phi(\vec{x}; \mathcal{W}))^{\gamma}\log(\Phi(\vec{x}; \mathcal{W})) \\
			& - (1 - y) \Phi(\vec{x}; \mathcal{W})^{\gamma} \log(1 - \Phi(\vec{x}; \mathcal{W})),
		\end{split}
	\end{equation*}
	where $\gamma$ is the focussing parameter that can be used to put increasing focus on misclassified examples \cite{Lin2020FocalLossDense}
}

\newdualentry{serr}{SEM}{standard error of the mean}{
	The variance of the mean's sampling distribution, defined as $\frac{\sigma}{\sqrt{n}}$, with the sample's standard deviation $\sigma$ and the sample's size $n$
}

\newglossaryentry{rgb}{name={RGB},description={
	Standard color model in images. RGB stands for \enquote{red, green, blue}, denoting the three color-channels whose respective luminances are encoded in the image
}}

\newglossaryentry{hsv}{name={HSV},description={
	Alternative color model in images, which better resembles the human perception of color. \Gls{hsv} stands for \enquote{hue, saturation, value}
}}

\newglossaryentry{adam}{name={Adam}, description={
	An alternative optimizer to \gls{sgd}, where the learning rate of each trainable paramter $w_i \in \mathcal{W}$ is adapted based on the current geometric curvature of $\frac{\delta L}{\delta w_t}$ \cite{Kingma2015AdamMethodStochastic}
}}

\newglossaryentry{radam}{name={RAdam}, description={
	Recitified Adam expands on Adam, and additionally rectifies the variance of the adaptive learning rate proposed in \gls{adam} \cite{Liu2020VarianceAdaptiveLearning}
}}

\newglossaryentry{led}{name={LED}, description={
	\Gls{led} stands for Light Emitting Diode, and is a special type of illuminant
}}

\newglossaryentry{umap}{name={UMAP}, description={
	\Gls{umap} is a technique that can be used for visualization similar to \gls{tsne}, but also for general, non-linear dimensionality reduction \cite{McInnes2018UmapUniformmanifold}.
	Compared to \gls{tsne}, \gls{umap} was shown to preserve the global structure of the original feature representation more accurately
}}
\makeglossaries
\glsdisablehyper
\robustify{\gls}
\def\mvt/{MVTec \gls{ad}}
\def\mvtl/{MVTec LOCO \gls{ad}}
\def\mvtt/{MVTec 3D-\gls{ad}}
\def\visa/{VisA}
\def\knn/{\textit{k}-\gls{nn}}

\begin{document}

\title{Optimizing PatchCore for Few/many-shot Anomaly Detection}

\author{João Santos, Triet Tran, Oliver Rippel\\
Scortex\\
22 Rue Berbier du Mets\\
{\tt\small jsantos@scortex.io}
}
\maketitle

\begin{abstract}
   Few-shot \gls{ad} is an emerging sub-field of general \gls{ad}, and tries to distinguish between normal and anomalous data using only few selected samples.
   While newly proposed few-shot \gls{ad} methods do compare against pre-existing algorithms developed for the full-shot domain as baselines, they do not dedicatedly optimize them for the few-shot setting.
   It thus remains unclear if the performance of such pre-existing algorithms can be further improved.
   We address said question in this work.
   Specifically, we present a study on the \gls{ad}/\gls{as} performance of PatchCore, the current state-of-the-art full-shot \gls{ad}/\gls{as} algorithm, in both the few-shot and the many-shot settings.
   We hypothesize that further performance improvements can be realized by (I) optimizing its various hyperparameters, and by (II) transferring techniques known to improve few-shot supervised learning to the \gls{ad} domain.
   Exhaustive experiments on the public \visa/ and \mvt/ datasets reveal that (I) significant performance improvements can be realized by optimizing hyperparameters such as the underlying feature extractor, and that (II) image-level augmentations can, but are not guaranteed, to improve performance.
   Based on these findings, we achieve a new state of the art in few-shot \gls{ad} on \visa/ (\cref{fig:patchcore-optimized_patchcore-default_winclip_comparison}), further demonstrating the merit of adapting pre-existing \gls{ad}/\gls{as} methods to the few-shot setting\footnote{We also obtained first place in the few-shot \gls{ad} track of the VAND challenge at CVPR 2023 with this approach, see \url{https://sites.google.com/view/vand-cvpr23/challenge}.}.
   Last, we identify the investigation of feature extractors with a strong inductive bias as a potential future research direction for (few-shot) \gls{ad}/\gls{as}.
\end{abstract}

\glsresetall

\section{Introduction}\label{sec:intro}%
Image \gls{ad} and \gls{as} are respectively tasked with identifying abnormal images, and with localizing the abormal patterns inside said images \cite{Chandola2009Anomalydetectionsurvey,Pimentel2014reviewnoveltydetection,Ruff2021UnifyingReviewDeep}.
Here, an anomaly is defined simply as being a deviation from a pre-defined/learned concept of normality \cite{Chandola2009Anomalydetectionsurvey,Pimentel2014reviewnoveltydetection,Ruff2021UnifyingReviewDeep}.
\Gls{ad}/\gls{as} play a crucial role in domains such as medical image analysis, video surveillance, or \gls{avi}, where anomalies correspond to diseases, threats, and defects, respectively \cite{Bergmann2019MVTecADComprehensive,Nazare2018Arepretrained,Schlegl2019fAnoGANFast}.
\begin{figure}[t]
   \centering
   \includegraphics[width=1.0\linewidth]{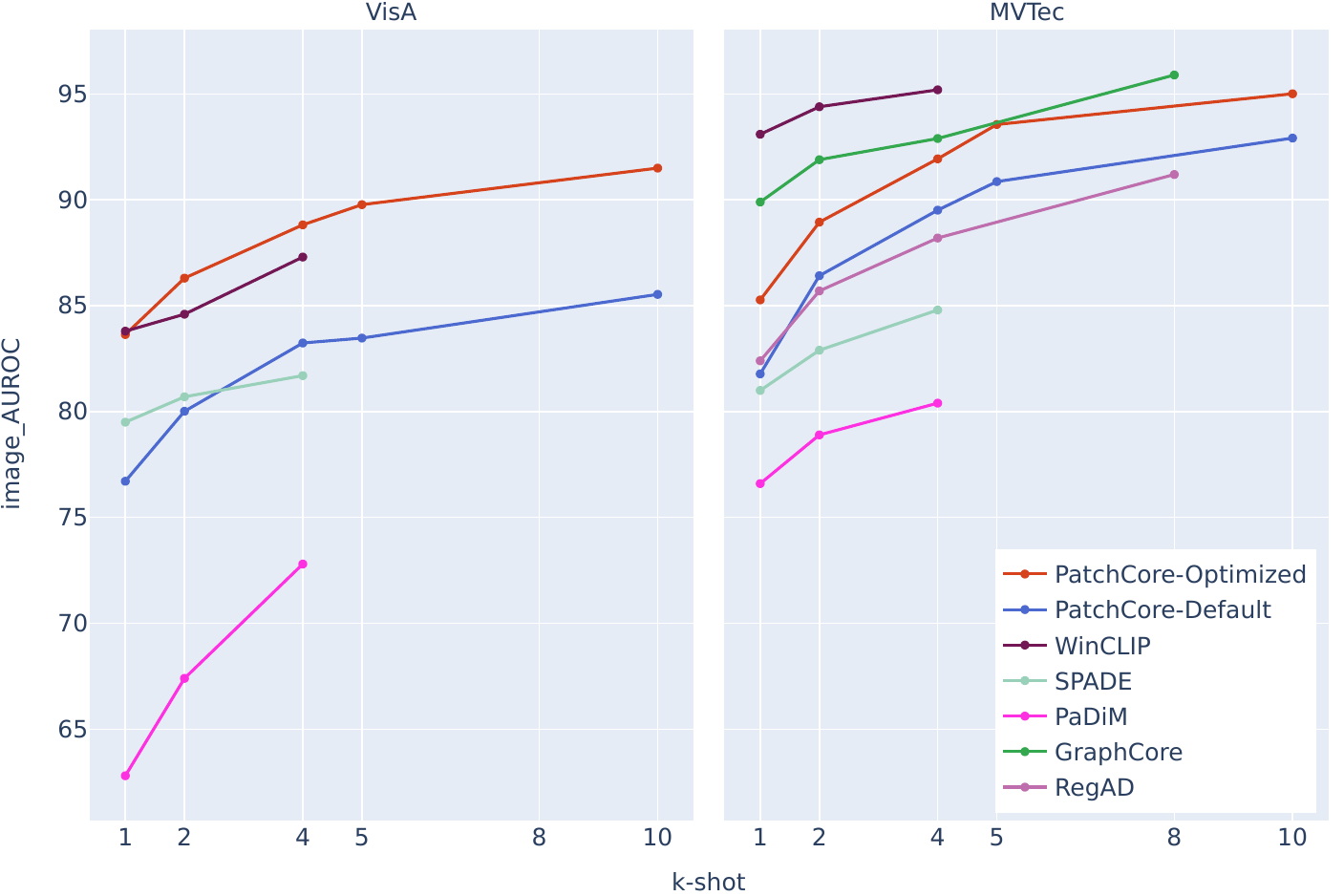}
   \caption{Optimizing the hyperparmeters of existing full-shot \gls{ad} methods such as PatchCore \cite{Roth2022TowardsTotalRecall} boosts their few-shot performance and furthermore yields state-of-the-art \gls{ad} results on \visa/ \cite{Zou2022SPotDifferenceSelf}.}
   \label{fig:patchcore-optimized_patchcore-default_winclip_comparison}
\end{figure}

Due to the limited availability of anomalous data, current \gls{ad}/\gls{as} methods are typically trained in a one-class fashion, using (large amounts of) normal data only \cite{Ruff2021UnifyingReviewDeep,Roth2022TowardsTotalRecall, Defard2021PaDiMPatchDistribution, Rippel2021GaussianAnomalyDetection, Cohen2020SubImageAnomaly,Zavrtanik2021DRAEMDiscriminativelyTrained,Yi2020PatchSVDDPatch,Rudolph2021SameSameDifferNet}.
They furthermore tend to exploit features from models pre-trained on large corpora of (natural) images such as ImageNet \cite{Deng2009ImageNetlargescale}, assuming that the target \gls{ad} domain will be close enough to the domain used for pre-training \cite{Roth2022TowardsTotalRecall, Defard2021PaDiMPatchDistribution, Rippel2021GaussianAnomalyDetection, Cohen2020SubImageAnomaly, Zavrtanik2020Reconstructioninpaintingvisual,Rudolph2021SameSameDifferNet}.
Some of these methods have shown outstanding performance on the full-shot setting, i.e.\ when training with \textgreater100 normal images.
As a consequence, popular datasets generally used for benchmarking developed methods are practically solved.
For example, the state-of-the-art method PatchCore \cite{Roth2022TowardsTotalRecall} obtains 99.6\% \gls{auroc} on \mvt/ \cite{Bergmann2019MVTecADComprehensive, Bergmann2021MVTecAnomalyDetection}, a dataset focused on \gls{avi}.

As a consequence to this saturation, the few-shot as well as the many-shot settings have been proposed recently, which aim to solve the \gls{ad}/\gls{as} task with less normal data \cite{Huang2022RegistrationbasedFew, Rudolph2021SameSameDifferNet,Sheynin2021hierarchicaltransformationdiscriminating,Roth2022TowardsTotalRecall}.
While no consensus has been established yet on where to draw the exact line between few-shot, many-shot, and full-shot \gls{ad}/\gls{as}, literature typically considers $\leq10$ images to be few-shot, and $\geq100$ images to be full-shot \cite{Huang2022RegistrationbasedFew,Rudolph2021SameSameDifferNet,Sheynin2021hierarchicaltransformationdiscriminating,Jeong2023WinCLIPZero/few,Aota2023ZeroShotVersus}.
Performing well in both the few and the many-shot setting is furthermore essential for domains where the acquisition of labeled normal data is very resource-intensive, such as the medical or industrial domains \cite{Schlegl2019fAnoGANFast,Roth2022TowardsTotalRecall}.
In both settings, performances comparable to the full-shot regime have furthermore not yet been achieved \cite{Huang2022RegistrationbasedFew,Rudolph2021SameSameDifferNet,Sheynin2021hierarchicaltransformationdiscriminating,Jeong2023WinCLIPZero/few,Aota2023ZeroShotVersus,Xie2023PushingLimitsFewshot}.
However, current efforts focus on developing novel algorithms for the few and the many-shot settings, rather than adapting/optimizing existing methods \cite{Huang2022RegistrationbasedFew,Jeong2023WinCLIPZero/few,Aota2023ZeroShotVersus,Xie2023PushingLimitsFewshot}.

In this work, we focus on improving existing \gls{ad}/\gls{as} algorithms in both the few-shot and the many-shot setting.
We argue that the hyperparameters of current \gls{ad}/\gls{as} methods are overfit to the full-shot setting, and that thus further performance gains can be realized by optimizing them.
We show this to be true for PatchCore \cite{Roth2022TowardsTotalRecall}, the current state-of-the-art full-shot \gls{ad}/\gls{as} algorithm.
We moreover investigate whether
data augmentation techniques, which are known to improve performance for supervised learning in the limited data regime, are also beneficial for few/many-shot \gls{ad}/\gls{as}.
Specifically, we show through extensive experiments, that:
\begin{itemize}
   \item The trend that architectures with stronger ImageNet performance are also better for \gls{ad}/\gls{as} \cite{Rippel2021GaussianAnomalyDetection} is not confirmed in the few and many-shot \gls{ad} regimes.
   \item Increasing the inductive bias of \glspl{cnn} by enforcing translation equivariance improves performance, revealing a potential avenue for future research.
   \item Image-level augmentations can improve few-shot performance overall, but must be tuned carefully to do so.
   \item Greedy coreset subsampling neither increases nor decrease performance, while simultaneously lowering the size of the memory bank significantly.
   \item Combining all of this leads to a new state of the art in few-shot \gls{ad} on the challenging \visa/ dataset \cite{Zou2022SPotDifferenceSelf} (\cref{fig:patchcore-optimized_patchcore-default_winclip_comparison}), demonstrating the merit of adapting pre-existing \gls{ad}/\gls{as} methods to the few/many-shot settings.
         This is further supported by the fact that we obtained first place in the few-shot \gls{ad} track of the VAND challenge at CVPR 2023 with this approach.
\end{itemize}
To ensure reproducibility, we provide the full code with detailed instructions at \url{https://github.com/scortexio/patchcore-few-shot/}.

\section{Related Work}

\subsection{\Glsentryshort{ad} in the Few/many-shot Regime}
Currently, only few works propose methods to specifically tackle the few/many-shot \gls{ad}/\gls{as} problem \cite{Huang2022RegistrationbasedFew, Jeong2023WinCLIPZero/few, Rudolph2021SameSameDifferNet,Xie2023PushingLimitsFewshot,Chen2023Zero/FewShot,Cao2023SegmentAnyAnomaly}.
Similarly to the full-shot regime, proposed methods generally use a pre-trained feature extractor to extract relevant features from the input image data \cite{Huang2022RegistrationbasedFew, Jeong2023WinCLIPZero/few, Rudolph2021SameSameDifferNet}.
However, there also exist methods that rely on learning features from scratch.
For example, authors in \cite{Xie2023PushingLimitsFewshot} propose learning visual, isometric invariant features with graph neural networks to handle the low amount of training data inherent to the few-shot setting.
Alternatively, vision-language based methods have been proposed recently \cite{Jeong2023WinCLIPZero/few,Cao2023SegmentAnyAnomaly,Chen2023Zero/FewShot}, which leverage pre-existing domain knowledge via language prompting and have shown impressive results both on the zero-shot and few-shot settings.
Vision-language-based methods are, however, quite expensive from a computational point of view, and thus not easily applicable under hardware/time constraints, present e.g.\ in \gls{avi}.
They moreover require careful prompt engineering to work, often leveraging pre-existing domain knowledge to maximize performances.

While recently developed few/many-shot \gls{ad}/\gls{as} methods compare among one another and report results of pre-existing full-shot methods applied to the few/many-shot setting as baselines \cite{Jeong2023WinCLIPZero/few,Xie2023PushingLimitsFewshot}, none of them dedicatedly optimize the baselines to the few/many-shot setting.
In other domains however, this was shown to be important, since baselines often demonstrated competitive performances when optimized to the target application/domain \cite{Oliver2018RealisticEvaluationDeep,Isensee2021nnUNetself}.

A baseline suitable for such optimizations is PatchCore \cite{Roth2022TowardsTotalRecall}, a state-of-the-art deep learning-based algorithm for image \gls{ad}/\gls{as}.
PatchCore stores locally aware patch features extracted from normal training images into a memory bank.
This memory bank represents a hyper-dimensional space of “normality”.
During inference, each patch from the inferred image is compared with all patches in the memory bank to obtain an anomaly score.
This score is expected to be higher for anomalous patches compared to normal ones.
PatchCore, intended to be used in the full-shot \gls{ad} regime, has been previously tested on the few/many-shot settings and has shown promising preliminary results here \cite{Roth2022TowardsTotalRecall, Jeong2023WinCLIPZero/few, Xie2023ImiadIndustrial,Xie2023PushingLimitsFewshot}.
However note that, as of now, PatchCore has not been dedicatedly optimized for these settings.
At most, authors of GraphCore \cite{Xie2023PushingLimitsFewshot} have combined PatchCore with image-level data augmentations to motivate their method.

\subsection{Transferring Techniques from Supervised Few-shot Learning to Anomaly Detection}
Techniques developed to improve supervised few-shot classification and/or transfer learning performance comprise data augmentation \cite{Perez2017effectivenessdataaugmentation, Cubuk2019AutoaugmentLearningaugmentation}, improved feature extractors \cite{Liu2022ConvNet2020s, Tan2019EfficientNetRethinkingModel, Dosovitskiy2021ImageisWorth}, and training on larger input sizes \cite{Tan2019EfficientNetRethinkingModel, Tan2021EfficientNetV2SmallerModels, Roth2022TowardsTotalRecall}, all of which have not been explored exhaustively in the few/many-shot \gls{ad} settings so far.
Moreover, the limited previous work that explores these techniques in context of few/many-shot \gls{ad}/\gls{as} \cite{Gutierrez2021Dataaugmentationpre, Zhang2023WhatMakesGood, Xie2023ImiadIndustrial} has focused their evaluations on simple datasets such as \mvt/ \cite{Bergmann2019MVTecADComprehensive,Bergmann2021MVTecAnomalyDetection}, rather than on more comprehensive and challenging datasets like \visa/ \cite{Zou2022SPotDifferenceSelf}.
The original authors of PatchCore \cite{Roth2022TowardsTotalRecall} furthermore analysed the performance of different resolutions and feature extractors, but only on the full-shot regime and only using \mvt/ \cite{Roth2022TowardsTotalRecall}.
Finally, authors in \cite{Heckler2023ExploringImportancePretrained} performed a comprehensive analysis on the influence of feature extractors, layer selection and input resolutions on different \gls{ad}/\gls{as} algorithms, but also limit their work to the full-shot \gls{ad} setting.
They further evaluate only on the \mvt/ dataset, and conducted their work concurrent to the study presented here.

Modifications like the ones mentioned above can be easily plugged into PatchCore, as detailed in the following.

\section{Optimizing PatchCore for Few/many-shot Anomaly Detection}

\subsection{Better Feature Extractors}
Since PatchCore is essentially a \knn/ search in intermediate feature representations of pre-trained feature extractors, performance gains should be possible by optimizing the used feature extractor.

Regarding feature extractors and their transferability, it has been shown that model architectures with better classification performance on large-scale and varied datasets, such as ImageNet \cite{Deng2009ImageNetlargescale}, also achieve better transfer learning performance on downstream tasks \cite{Kornblith2019Dobetterimagenet}.
Note that literature is currently inconclusive on whether this trend also holds for methods that leverage pre-trained feature extractors for \gls{ad} \cite{Rippel2021GaussianAnomalyDetection,Heckler2023ExploringImportancePretrained}.
Moreover, models with a weak inductive bias, e.g.\ transformer-based architectures, have been introduced in the computer vision domain recently \cite{Dosovitskiy2021ImageisWorth}.
These models have achieved state-of-the-art results, both on the full-shot \cite{Dosovitskiy2021ImageisWorth}, but also on the few-shot supervised learning settings \cite{Hu2022Pushinglimitssimple}.
From this, it can be hypothesized that less-constrained models would also perform well in the few/many-shot \gls{ad}/\gls{as} setting.

Contrary to this, a stronger inductive bias is generally assumed to require less data to perform well \cite{James2014IntroductionStatisticalLearning,Belkin2019Reconcilingmodernmachine}, since some assumptions regarding the underlying data are already incorporated into the model itself.
Those assumptions thus do not need to be learned from the data by the model.
We explore more constrained models in our work, as we believe them to be crucial for performing well in the few-shot setting, where the low amount of training data increases the risk of mismatch between training and real data distributions.
Specifically, we investigate how much anti-aliasing, a technique used to restore the translation equivariance of \glspl{cnn} \cite{Zhang2019Makingconvolutionalnetworks}, will affect few/many-shot \gls{ad}/\gls{as} performance.

\subsection{Increased Resolution}
Typically, there is a trade-off between the computational cost and the size of the input image in the sense that a higher input resolution allows us to capture more fine-grained details at the cost of a bigger computational cost, and potentially more unstable training \cite{Touvron2019Fixingtraintest}.

In the full-shot \gls{ad} regime, increasing the resolution of the input images to extract finer-grained features has shown to boost \gls{ad}/\gls{as} performance \cite{Roth2022TowardsTotalRecall}.
While there is additional research needed to identify the underlying mechanisms here, initial work hypothesizes that the detectability of defects is determined by the combination of their size and the effective receptive field sizes of the pre-trained feature extractor's layers \cite{Heckler2023ExploringImportancePretrained}.
Integrating this into PatchCore is straightforward for \glspl{cnn}-based feature extractors, and possible with tricks for transformer-based networks \cite{Dosovitskiy2021ImageisWorth,Liu2021SwintransformerHierarchical}.

\subsection{Data Augmentation}

Data augmentation tries to improve the generalization of models by adding sensible variability to the training dataset \cite{Perez2017effectivenessdataaugmentation, Cubuk2019AutoaugmentLearningaugmentation}.
It has been extensively used in the supervised learning domain on different data types, ranging from images, text to audio \cite{Perez2017effectivenessdataaugmentation, Wei2020ComparisonDataAugmentation, Bayer2022surveydataaugmentation}.
Some studies have also looked into performing a search for the best augmentation setup for each dataset \cite{Cubuk2019AutoaugmentLearningaugmentation}.
Related work has further shown for specific data augmentations such as rotations or translations to improve \gls{ad} performance \cite{Xie2023ImiadIndustrial, Gutierrez2021Dataaugmentationpre}, and we therefore include them in the considered/assessed augmentations.

Conversely, other techniques have been developed that augment the intermediate feature representations of the model directly \cite{Wang2021RegularizingDeepNetworks,Ouali2020SemiSupervisedSemantic}, instead of modifying the input image space.
These variations are designed to induce semantically meaningful changes to the feature space, and thus require assumptions about the semantics of the feature space (e.g.\ target classes should follow Gaussian distributions in them \cite{Wang2021RegularizingDeepNetworks}).
Conceptually, feature-based augmentations are much more elegant than image-based augmentations, since they require less expertise in the source domain, and the same augmentations could potentially be applied across domains.
While we did investigate feature-based augmentations, we did not succeed.
We thus leave their study to future work.

Both image-level and feature-level augmentations can be easily integrated into PatchCore during the construction of the \knn/ memory bank (i.e.\ by augmenting either input images prior to feature extraction or by augmenting the extracted feature representations).

\subsection{Coreset Subsampling}
Coreset selection is a technique used in machine learning and data analysis to identify a representative subset of data that maintains the same topology as the full dataset \cite{Agarwal2004Geometricapproximationvia,Naik2022FastBayesianCoresets}.
PatchCore uses this technique to significantly reduce its memory bank size, such that the inference time and memory usage can be kept low when we have a higher amount of training images, input resolution, or embedding dimensionality.

PatchCore was shown to work well in conjunction with a greedy coreset construction, which can also slightly improve performance in the full-shot \gls{ad} regime \cite{Roth2022TowardsTotalRecall}.
In the few-shot regime this technique has not been extensively explored, possibly since the memory bank is already small here.
We hypothesize that this technique may also bring a benefit to the few-shot/many-shot \gls{ad} settings if used in conjunction with data augmentation techniques, and thus investigate its use in this work.

\section{Experiments and Results}

\subsection{General Setup}
We select PatchCore for optimization due to its outstanding full-shot \gls{ad}/\gls{as} performance combined with its simplicity, and evalute on \mvt/ \cite{Bergmann2019MVTecADComprehensive,Bergmann2021MVTecAnomalyDetection} and \visa/ \cite{Zou2022SPotDifferenceSelf} to evaluate performances.
\mvt/ \cite{Bergmann2019MVTecADComprehensive,Bergmann2021MVTecAnomalyDetection} and \visa/ \cite{Zou2022SPotDifferenceSelf} constitute the two biggest publicly available \gls{ad}/\gls{as} datasets dealing with \gls{avi}.
In total, they comprise 22 objects and 5 texture categories, with \textgreater \num{2400} anomalous and \textgreater \num{13500} normal images.

We evaluate in the few-shot (1,5 and 10) and in the many-shot (25 and 50) settings, and draw the training samples from each respective category's training set.
For each object category, every k-shot was run with randomly chosen training seeds/images for a total of 5 times, and the same seeds/images are used across all experiments to ensure direct comparability.
We furthermore used the PatchCore implementation from the anomalib library as the base \cite{Akcay2022AnomalibDeepLearning}, and made modifications according to our experiments.
We used the default configuration of anomalib, but fixed the number of \glspl{nn} used for calculating the anomaly score to~1.
We observed a small improvement here compared to the re-weighting of scores from multiple neighbors proposed by PatchCore, but did not ablate on this.

\subsection{Metrics}
To evaluate image-wise \gls{ad} performance, we computed the \gls{auroc} \cite{Provost1997Analysisvisualizationclassifier}.
We followed \cite{Zou2022SPotDifferenceSelf} and computed the \gls{aupr} \cite{Davis2006relationshipPrecisionRecall} to evaluate \gls{as} performance due to the extreme class imbalance on the pixel-level.
Finally, to evaluate the joint \gls{ad}/\gls{as} performance, we used the Harmonic mean of image-\gls{auroc} and pixel-\gls{aupr} (HPROC), arguing that consistent performance across both \gls{ad}/\gls{as} tasks is important.
We have not used the F1-max score as done in previous work \cite{Zou2022SPotDifferenceSelf, Jeong2023WinCLIPZero/few}, since the \gls{pr} curve, and consequently the F1-max score derived from it, is overestimated when we have more anomalous than normal samples.
This is the case for the test sets of both \mvt/ and \visa/ at the image-level.
To analyse the overall performance in the few/many-shot domains, we propose to plot the HPROC values vs.\ the respective k-shots, and compute the area under this HPROC-k-shot curve (AUHPROC).

\subsection{Influence of Dataset, Feature Extractor and Input Resolution}
We start by analysing the influence of dataset, employed feature extractors, and input resolution on \gls{ad}/\gls{as} performances in the few and many-shot regimes.
To keep the computational cost at a reasonable level, we perform neither coreset subsampling nor image-level augmentations.

\textbf{Feature extractors:}
We assess a set of different feature extractors.
For each feature extractor, we do not ablate over the layer selection used, but instead we follow previous work by using intermediate to later layers of the network \cite{Roth2022TowardsTotalRecall, Heckler2023ExploringImportancePretrained}.
We note, however, that jointly optimizing this might yield further performance improvements \cite{Heckler2023ExploringImportancePretrained}.
Concretely, we use the following combination of feature extractors, original input resolution and layer selection\footnote{For complete details please check the code available at \url{https://github.com/scortexio/patchcore-few-shot/}}:
\begin{itemize}
   \item WideResNet50 \cite{Zagoruyko2016WideResidualNetworks} (224x224, layers 2 and 3), since it is the baseline used by the majority of \gls{ad}/\gls{as} methods and the default backbone used in PatchCore.
   \item EfficientNetB4 \cite{Tan2019EfficientNetRethinkingModel} (380x380, layers 4, 6 and 7) and ConvNext Small/Base \cite{Liu2022ConvNet2020s} (384x384, layers stages.2.blocks.10 and stages.2.blocks.20), to test if better ImageNet performance, which correlates with better transfer learning performance (i.e.
         more universal feature descriptors), also correlates with better few-shot \gls{ad}/\gls{as} performance.
         In addition, two ConvNext variants are used to investigate the influence of complexity \cite{Rippel2021ModelingDistributionNormal, Defard2021PaDiMPatchDistribution}.
   \item Anti-aliased WideResNet50 \cite{Zhang2019Makingconvolutionalnetworks} (224x224, layers 2 and 3), in order to see if a stronger inductive bias is useful for few/many-shot \gls{ad}/\gls{as}.
\end{itemize}

\textbf{Input size ranges:}
We benchmarked each feature extractor on a range of scales corresponding to 0.5x, 1x, 1.5x and 2x the original input resolution of each model.
The assumption here is that higher resolutions increase performance similar as for supervised learning, as we remove less information when downscaling.
The 0.5x scale is used to further verify said assumption.

\textbf{Results:}
In order to better understand each of the previous factors, we present the respective performances obtained after marginalizing over the other two parameters.

To analyse the influence of the dataset on the \gls{ad}/\gls{as} performance, we report the mean and standard deviation of category-wise AUHPROC values computed over the few/many-shot settings after averaging over the 5 seeds.
\begin{table}[tb]
   \centering
   \caption{Performance comparison between \mvt/ and \visa/ on few-shot and many-shot regimes.
      We report the mean and standard deviation of category-wise AUHPROC values computed after averaging over seed/runs.}
   \label{table:dataset_comparison}
   \begin{tabular}{lrrrr}
      \toprule
              & \multicolumn{4}{c}{AUHPROC}                                                \\
      \cmidrule(lr){2-5}
              & \multicolumn{2}{c}{few-shot} & \multicolumn{2}{c}{many-shot}               \\
      \cmidrule(lr){2-3}
      \cmidrule(lr){4-5}
      dataset & mean                         & std                           & mean & std  \\
      \midrule
      \mvt/   & 58.8                         & 18.0                          & 59.3 & 17.9 \\
      \visa/  & 42.6                         & 22.7                          & 43.1 & 22.9 \\
      \bottomrule
   \end{tabular}
\end{table}

\Cref{table:dataset_comparison} shows that, on average, the performance on \mvt/ is higher than on \visa/ in both few and many-shot settings.
Moreover, the performance on the categories contained in \visa/ varies more than the performance on the categories contained in \mvt/.
Overall, this seems to indicate that, across all feature extractors and scales, the \visa/ dataset is more challenging than \mvt/ also in the few/many-shot settings.
We hypothesize the defect size to be one of the key differentiators between \mvt/ and \visa/, and report the distribution of defect sizes in \cref{appendix_defect_size}.
Therefore, future research in few/many-shot \gls{ad}/\gls{as} should focus on more challenging datasets such as \visa/ or \mvtl/ \cite{Bergmann2022DentsScratchesLogical}, rather than evaluating solely on \mvt/.

\begin{figure}[t]
   \centering
   \includegraphics[width=1.0\linewidth]{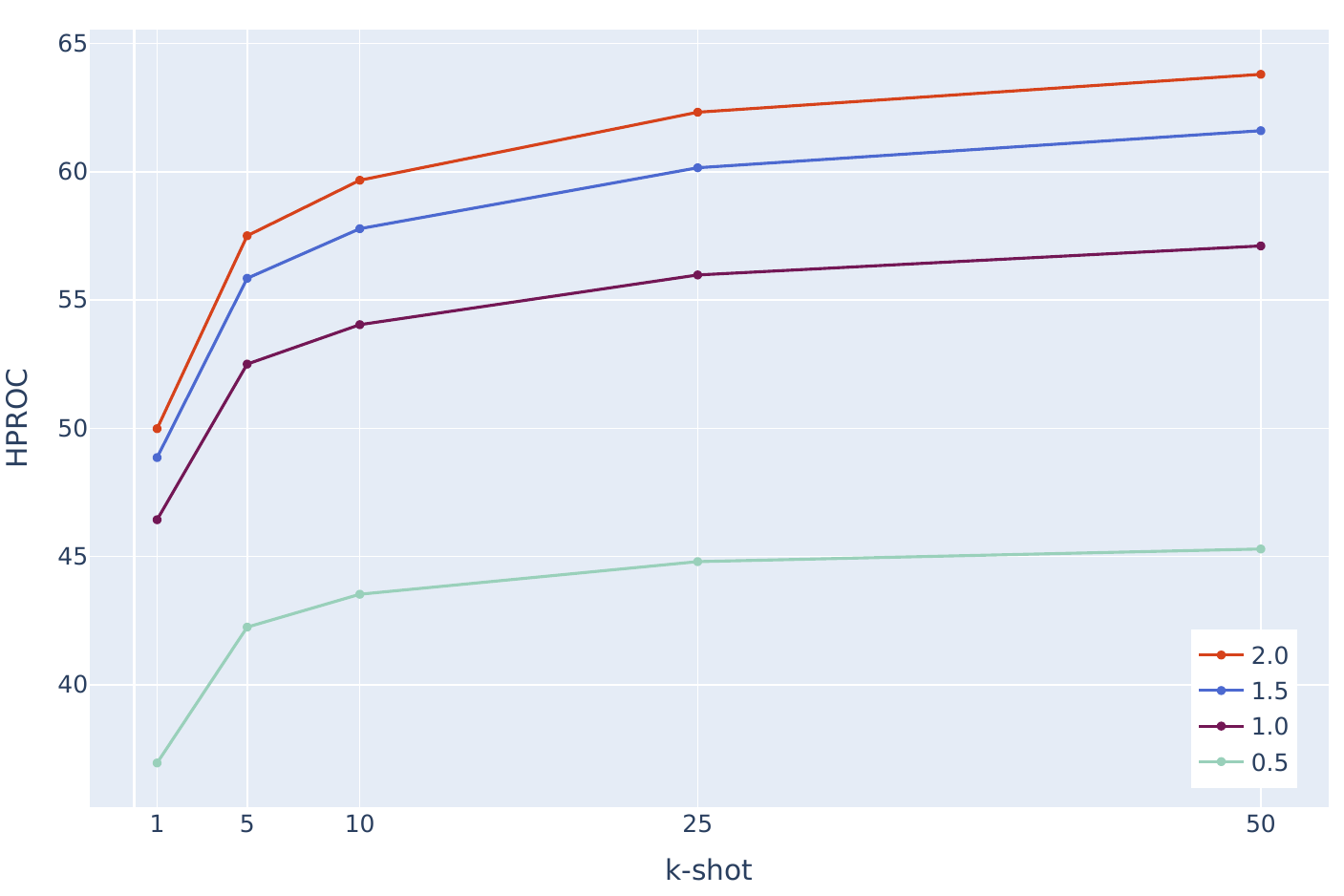}
   \caption{HPROC across k-shots for different input scales, averaged over feature extractors, datasets and seeds/runs.}
   \label{fig:resolutions}
\end{figure}

To analyse the effect of the scale on performance, we plot in \cref{fig:resolutions} the HPROC-k-shot curve for different input scales, averaged over feature extractors, datasets and seeds/runs.
An increase in scale generally helps across datasets and feature extractors not only in the full-shot setting as indicated in \cite{Roth2022TowardsTotalRecall}, but also in the few/many-shot settings.
Furthermore, it would be interesting to see if the underlying reasons for the performance improvements are truly just that an increase in scale effectively "shifts" the detectability of the defects to be more in the selected layers by modulating the effective receptive field size, as proposed recently \cite{Heckler2023ExploringImportancePretrained}.
This hypothesis moreover runs counter to the intuition that later layers of a \gls{cnn}/model encode more semantically abstract features compared to earlier layers \cite{Yosinski2015Understandingneuralnetworks}.
We believe that performing a fundamental validation of said hypothesis, for both semantic/logical as well as structural/textural anomaly types \cite{Bergmann2022DentsScratchesLogical,Ruff2021UnifyingReviewDeep}, is an interesting avenue of future work.

Finally, to assess the performance of different feature extractors, we compute the AUHPROC category-wise after averaging over the 5 seeds.
We then compute mean as well as standard deviation of category-wise AUHPROC for each scale/feature extractor combination.
HPROC vs. k-shot curves aggregated over all scales are moreover reported in \cref{appendix_results_featureextractor-vs-scale}.
\begin{table}[t]
   \centering
   \caption{Performance comparison between different backbones at different input scales.
      We report mean and standard deviation of category-wise AUHPROC generated after averaging over seeds/runs.
      Best/second best value per scale are boldfaced/underlined respectively.
      Abbreviations:
      AA~=~Anti-aliased WideResNet50,
      WRN~=~WideResNet50,
      CN-B~=~ConvNext Base,
      CN-S~=~ConvNext Small,
      ENB4~=~EfficientNetB4.}
   \label{table:backbone_scale_comparison}
   \renewcommand{\tabcolsep}{3pt}
   \begin{tabular}{@{}lSSSSSSSS@{}}
      \toprule
           & \multicolumn{8}{c}{AUHPROC}                                                                                                                                                      \\
      \cmidrule(lr){2-9}
           & \multicolumn{2}{c}{0.5x}    & \multicolumn{2}{c}{1.0x} & \multicolumn{2}{c}{1.5x} & \multicolumn{2}{c}{2.0x}                                                                     \\
      \cmidrule(lr){2-3}
      \cmidrule(lr){4-5}
      \cmidrule(lr){6-7}
      \cmidrule(lr){8-9}
           & {mean}                      & {std}                    & {mean}                   & {std}                    & {mean}         & {std}          & {mean}         & {std}          \\
      \midrule
      AA   & 44.1                        & 29.5                     & \bfseries 57.9           & 26.8                     & \bfseries 63.1 & 22.1           & \bfseries 66.2 & 18.2           \\
      WRN  & 43.5                        & 29.1                     & \Uline{56.6}             & 25.8                     & \Uline{62.0}   & 21.9           & \Uline{64.9}   & 19.1           \\
      CN-B & \bfseries 46.6              & \Uline{26.3}             & 56.1                     & \Uline{21.2}             & 59.5           & \Uline{19.1}   & 60.5           & \Uline{17.1}   \\
      CN-S & \Uline{46.3}                & \bfseries 24.6           & 54.9                     & \bfseries 19.6           & 57.4           & \bfseries 17.4 & 58.3           & \bfseries 15.3 \\
      ENB4 & 40.9                        & 28.5                     & 51.5                     & 25.5                     & 55.3           & 22.5           & 57.6           & 20.3           \\
      \bottomrule
   \end{tabular}
\end{table}
\Cref{table:backbone_scale_comparison} shows that Anti-aliased WideResNet50 performs best on average compared to the other backbones, followed by the WideResNet50.
However, high standard deviations are observed for both versions of the WideResNet50 at lower input scales, and ConvNext models seem to be more consistent across shots and scales overall.
Last, the EfficientNetB4 model is neither as capable nor as consistent as any other model that was evaluated.
While all assessed feature extractors benefit from an increase in input scale, the realized performance improvement varies across feature extractors.
For example, the WideResNet50 improves average AUHPROC by \num{8.3} while reducing standard deviation by \num{11.3} when increasing input scale from 1x to 2x, whereas ConvNext Base improves average AUHPROC only by \num{4.4} while reducing standard deviation by \num{4.1}.
We thus hypothesize that additional variation was incurred for WideResNet-type models by the combination of their low initial input resolution and small defect sizes (see also \cref{appendix_results_scale-vs-datset}).

Overall, results presented here do not confirm the positive correlation between ImageNet performance and transfer learning performance \cite{Kornblith2019Dobetterimagenet} for few/many-shot \gls{ad}/\gls{as}, as one would expect ConvNext to be best performing then, followed by EfficientNet.
Instead, both WideResNet50 and its anti-aliased version outperform both ConvNext and EfficientNet.
Moreover, it can be seen that the "best" feature extractor is different per assessed category (see \cref{appendix_results_featureextractor-vs-datset}), which is in line with related work \cite{Heckler2023ExploringImportancePretrained}.
Furthermore, models of medium complexity remain sufficient for few/many-shot \gls{ad}/\gls{as} performance, since both versions of ConvNext perform comparably.
In all, the question "What makes a good feature extractor for (few/many-shot) \gls{ad}/\gls{as}?" therefore still remains unanswered at large.

Our experiments indicate, however, that more constrained feature extractors might be a valid avenue of future research, as the anti-aliased feature extractor outperformed all others.
This line of research is orthogonal to approaches that incorporate these constraint into the \gls{ad}/\gls{as} methods, either directly or indirectly \cite{Xie2023PushingLimitsFewshot,Artola2023GLADGlobalLocal,Huang2022RegistrationbasedFew}.
As a next step, one should perform an in-depth comparison of more constrained feature extractors here, and furthermore include rotation-equivariant feature extractors \cite{Weiler2018LearningSteerableFilters,Cesa2022ProgramBuildEN} into said comparison (see \cref{subsec:per_category_results} for a motivation).

\subsection{Data Augmentation}
For this set of experiments, we set out to understand whether we could improve performance by augmenting our training set.
We therefore artificially increased our training datasets by applying different sets of augmentations, implemented in the albumentations library \cite{Buslaev2020AlbumentationsFastFlexible}:
(I) Affine transformations (rotation, scale and translation),
(II) Blur,
(III) RandomBrighnessContrast changes,
(IV) Sharpening, and
(V) Flipping (horizontal and vertical).
Augmentations were carefully selected considering both prior work \cite{Xie2023ImiadIndustrial, Gutierrez2021Dataaugmentationpre} and the evaluated dataset, since flipping constitutes an anomaly for some categories of \mvt/ (e.g.\ transistor) \cite{Rippel2021ModelingDistributionNormal}.
We therefore used the five augmentations outlined above for the \visa/ dataset, while restricting ourselves to the first four for the \mvt/ dataset.
For additional details about the augmentations' hyperparameters we refer to our code.

To now analyse the effect of the data augmentations, we disabled each respective augmentation in an ablation fashion, while keeping the remaining ones active.
The number of augmentations was fixed across k-shots, and we augmented each training image 8 times, with only one type of augmentation being applied per augmented image.
The augmentation applied per image was furthermore randomly drawn from the set of all augmentations.
Evaluations are moreover restricted to the few-shot regime (i.e.\ 1, 5 and 10 shots) due to VRAM limitations.
We further constrained the feature extractor to the Anti-aliased WideResNet50 and the input scale to 2x, given that this hyperparameter combination provided the best performance in the previous set of tests and to reduce the computational cost of the evaluations.

\textbf{Results:}
One can see in \cref{table:augmentation_comparison} that data augmentation can actually decrease performance on \mvt/.
More specifically, we see that Sharpen seems to have an adverse effect on the achieved performance, since only its removal increases performance beyond the baseline.
Repeating the experiment for the \mvt/ dataset under the exclusion of the Sharpen augmentation further shows that only the combination of all augmentations outperforms the baseline (see \cref{appendix_influence_aug}).
On the contrary, for the \visa/ dataset, all assessed augmentation combinations outperform the baseline.
Removing the Flip augmentation noticeably decreases the performance gain by a large margin, indicating that this augmentation is quite important for this dataset.
We see that Affine, Blur and Sharpen do not seem to improve the performance of the system, given that the AUHPROC increases when these augmentations are removed.
Furthermore, it is worth noting that BrightnessContrast augmentation does not seem to impact performance at all.

\begin{table}[t]
   \centering
   \caption{Influence of augmentations on few-shot performance.
      We disabled each respective augmentation while keeping the remaining ones active.
      We report the mean of category-wise AUHPROC generated after averaging over seeds/runs.
      We also report the relative gain/loss compared to the baseline of not augmenting.
      Best value per dataset are boldfaced.
      Abbreviations:
      \#~augs~=~number of augmentations,
      A~=~Affine,
      BC~=~BrightnessContrast,
      B~=~Blur,
      S~=~Sharpen,
      F~=~Flip.
   }
   \begin{tabular}{lllllllrr}
      \toprule
                                                        &                       & \multicolumn{5}{c}{augmentation type} & \multicolumn{2}{c}{AUHPROC}                                                                                            \\
      \cmidrule(lr){3-7}
      \cmidrule(l){8-9}
                                                        & \# augs               & A                                     & BC                          & B                          & S                          & F      & mean           & diff \\
      \midrule
      \multirow{6}{*}{\rotatebox[origin=c]{90}{\mvt/}}  & 0                     & \xmark                                & \xmark                      & \xmark                     & \xmark                     & \xmark & 67.6           & 0.0  \\
      \cmidrule(l){2-9}
                                                        & \multirow[t]{5}{*}{8} & \xmark                                & \cmark                      & \cmark                     & \cmark                     & \xmark & 67.1           & -0.5 \\
                                                        &                       & \multirow[t]{4}{*}{\cmark}            & \xmark                      & \cmark                     & \cmark                     & \xmark & 67.0           & -0.6 \\
                                                        &                       &                                       & \multirow[t]{3}{*}{\cmark}  & \xmark                     & \cmark                     & \xmark & 67.0           & -0.6 \\
                                                        &                       &                                       &                             & \multirow[t]{2}{*}{\cmark} & \xmark                     & \xmark & \bfseries 68.2 & 0.6  \\
                                                        &                       &                                       &                             &                            & \cmark                     & \xmark & 66.8           & -0.8 \\
      \cmidrule(l){1-9}
      \multirow{7}{*}{\rotatebox[origin=c]{90}{\visa/}} & 0                     & \xmark                                & \xmark                      & \xmark                     & \xmark                     & \xmark & 52.5           & 0.0  \\
      \cmidrule(l){2-9}
                                                        & \multirow[t]{6}{*}{8} & \xmark                                & \cmark                      & \cmark                     & \cmark                     & \cmark & \bfseries 56.0 & 3.5  \\

                                                        &                       & \multirow[t]{5}{*}{\cmark}            & \xmark                      & \cmark                     & \cmark                     & \cmark & 55.0           & 2.5  \\

                                                        &                       &                                       & \multirow[t]{4}{*}{\cmark}  & \xmark                     & \cmark                     & \cmark & 55.8           & 3.3  \\

                                                        &                       &                                       &                             & \multirow[t]{3}{*}{\cmark} & \xmark                     & \cmark & 55.5           & 3.0  \\

                                                        &                       &                                       &                             &                            & \multirow[t]{2}{*}{\cmark} & \xmark & 53.7           & 1.2  \\
                                                        &                       &                                       &                             &                            &                            & \cmark & 55.0           & 2.5  \\

      \bottomrule
   \end{tabular}
   \label{table:augmentation_comparison}
\end{table}

Overall, it seems that there is a benefit to be gained by using augmentations in the few to many-shot \gls{ad}/\gls{as} settings.
However, one is required to specifically tune the augmentations for each dataset to avoid accidentally decreasing performance.
Note that this observation is in agreement with literature \cite{Zhang2023WhatMakesGood}.
Moreover, performance boosts are much less pronounced compared to optimizing either feature extractor or input scale (see \cref{table:backbone_scale_comparison}), and vary strongly between datasets (gain of 0.035 AUHPROC for \visa/ compared to 0.006 AUHPROC for \mvt/).
We thus suggest that the former are optimized before investigating data augmentation techniques, and to employ data augmentation with care.

\subsection{Coreset Subsampling and Data Augmentation}
In the few-shot setting, using coreset in conjunction with data augmentation techniques might allow us to maintain performance improvements of the data augmentations while keeping the inference time low.
We explore this possibility by applying a fixed number of augmentations to each k-shot while varying the coreset size.
Specifically, we test applying 8 and 16 augmentations per training image, and vary the coreset size to be equal to that of 1, 4 or 8 images (corresponding to \num{3136}, \num{12544} and \num{25088} points in the memory bank, respectively).
We activated all the valid augmentations for the \visa/ dataset and the three augmentations for \mvt/ that provided a benefit on the previous tests (Blur, BrightnessContrast and Affine).
For direct comparability, we again used the Anti-aliased WideResNet50 as the feature extractor at 2x input scale.
We furthermore again limit ourselves to the few-shot setting due to VRAM limitations.

\textbf{Results:}
\begin{table}[t]
   \centering
   \caption{Combining coreset subsampling with augmentations for few-shot \gls{ad}/\gls{as}.
      We report mean and standard deviation of category-wise AUHPROC after averaging over seeds/runs.
      We use the Anti-aliased WideResNet50 as a feature extractor with an input scale of 2x.
      Best values per dataset are boldfaced.}
   \label{table:coreset_results}
   \begin{tabular}{lllrr}
      \toprule
                                                        &                                 &                  & \multicolumn{2}{c}{AUHPROC}                  \\
      \cmidrule(lr){4-5}
                                                        & coreset size                    & \# augmentations & mean                        & std            \\
      \midrule
      \multirow{8}{*}{\rotatebox[origin=c]{90}{\mvt/}}  & \multirow[t]{2}{*}{all}         & 0                & 67.6                        & 21.2           \\
                                                        &                                 & 8                & \bfseries 68.2              & \bfseries 20.3 \\
      \cmidrule(lr){2-5}
                                                        & \multirow[t]{2}{*}{\num{3136}}  & 8                & 68.1                        & 20.4           \\
                                                        &                                 & 16               & 67.3                        & 21.8           \\
      \cmidrule(lr){2-5}
                                                        & \multirow[t]{2}{*}{\num{12544}} & 8                & \bfseries 68.2              & \bfseries 20.3 \\
                                                        &                                 & 16               & 67.4                        & 21.5           \\
      \cmidrule(lr){2-5}
                                                        & \multirow[t]{2}{*}{\num{25088}} & 8                & \bfseries 68.2              & \bfseries 20.3 \\
                                                        &                                 & 16               & 67.4                        & 21.5           \\
      \midrule

      \multirow{8}{*}{\rotatebox[origin=c]{90}{\visa/}} & \multirow[t]{2}{*}{all}         & 0                & 52.5                        & \bfseries 20.0 \\
                                                        &                                 & 8                & 55.0                        & 20.8           \\
      \cmidrule(lr){2-5}
                                                        & \multirow[t]{2}{*}{\num{3136}}  & 8                & 54.9                        & 21.2           \\
                                                        &                                 & 16               & 54.6                        & 20.5           \\
      \cmidrule(lr){2-5}
                                                        & \multirow[t]{2}{*}{\num{12544}} & 8                & \bfseries 55.1              & 21.3           \\
                                                        &                                 & 16               & 54.7                        & 20.5           \\
      \cmidrule(lr){2-5}
                                                        & \multirow[t]{2}{*}{\num{25088}} & 8                & \bfseries 55.1              & 21.3           \\
                                                        &                                 & 16               & 54.7                        & 20.5           \\
      \bottomrule
   \end{tabular}
\end{table}
\Cref{table:coreset_results} shows that small coreset sizes are already sufficient to fully recapitulate the topology of the augmented training data.
Specifically, complete recapitulation is realized at 4 images (\num{12544} points), since further increasing the coreset size does not further improve achieved performance.
Moreover, further scaling the augmentation rate from 8 to 16 augmentations per image actually degrades achieved performance.
This could indicate that there is some level of uncertainty/instability during greedy coreset selection when the ratio of true data to augmented data becomes too small.
Alternatively, augmenting too much might skew the underlying training data distribution/topology.

Overall, coreset subsampling can be applied with a target coreset size of a few images, while augmenting the dataset, without decreasing the performance.
Alternative coreset selection approaches such as bayesian coreset subsampling \cite{Naik2022FastBayesianCoresets} might be investigated for a more stable as well as potentially faster coreset selection process in future work.

\subsection{Comparison with the State of the Art}\label{sec:sota}
We continue by comparing the \gls{ad} performance of PatchCore optimized for the few/many-shot regime with state-of-the-art few-shot \gls{ad} methods.
We focus on \gls{ad} performance rather than on joint \gls{ad}/\gls{as} performance to compare with as many methods as possible:
While not all \gls{ad} methods can be used for \gls{as}, \gls{ad} outputs can be readily constructed for any \gls{as} method.
Moreover, we evaluate on $k \in \{1,2,4\}$, since we found that most developed methods report \gls{ad} performance for this set of k.
To denote overall \gls{ad} performance, we computed the arithmetic mean of the respective statistics over all values of k, i.e.\ weighting each shot equally.
The optimal hyperparameters of PatchCore for the few-shot regime are as follows:
(I) Feature extractor: Anti-aliased WideResNet50,
(II) Input image size: 448$\times$448,
(III) Number of augmentations per image: 8,
(IV) Coreset size: \num{12544}.

\textbf{Results:}
The optimized version of PatchCore depends the least on the samples selected for training on both \mvt/ and \visa/, as denoted by its low standard deviations (\cref{table:state_of_the_art_comparison}).
It moreover demonstrates favorable results on \mvt/, achieving third place, and furthermore sets a new state of the art in \gls{ad} on the more challenging \visa/ dataset, where it is both the most consistent and most performant method.
Also, both WinCLIP and GraphCore do not benefit as much from additional training data, as denoted by smaller increases in image-\gls{auroc} when increasing k (see also \cref{fig:patchcore-optimized_patchcore-default_winclip_comparison}).
For WinCLIP, this can be explained by the strong influence of language-guidance on the overall performance, as denoted by its impressive zero-shot results on \mvt/ \cite{Jeong2023WinCLIPZero/few}.
Language-guidance is furthermore not used by WinCLIP currently when incorporating available training images in few/many-shot \gls{ad}, and we believe this to be an avenue of future research.
For GraphCore \cite{Xie2023PushingLimitsFewshot}, we argue this is due to the strong biases incorporated in learning geometrically invariant feature representations.
As a consequence, more training images bring only limited value, since only non-geometrical changes will be leveraged/learned by GraphCore.

Overall, this comparison validates our work, affirming that there is value in adapting pre-existing methods to few/many-shot \gls{ad}/\gls{as}.
Still, optimized Patchcore fails at times, and we provide an analysis of its failure cases in \cref{subsec:per_category_results}.

\begin{table}[t]
   \centering
   \caption{Comparison with the state the art.
      We report the mean and standard deviation of \gls{ad} performance over 5 random seeds each.
      Values for SPADE \cite{Cohen2020SubImageAnomaly}, PaDiM \cite{Defard2021PaDiMPatchDistribution}, and WinCLIP \cite{Jeong2023WinCLIPZero/few} are sourced from the WinCLIP paper \cite{Jeong2023WinCLIPZero/few}, whereas values for GraphCore \cite{Xie2023PushingLimitsFewshot} and RegAD \cite{Huang2022RegistrationbasedFew} are sourced from the respective papers.
      Best/second best value per dataset and k-shot are boldfaced/underlined respectively.
      "PCore ori" corresponds to anomalib's \cite{Akcay2022AnomalibDeepLearning} reimplementation of the PatchCore algorithm, using the default settings, whereas "PCore opt" corresponds to its optimized version.}
   \label{table:state_of_the_art_comparison}
   \renewcommand{\tabcolsep}{2.7pt}
   \begin{tabular}{@{}llSS[table-format=1.1]SS[table-format=1.1]SS[table-format=1.1]SS[table-format=1.1]@{}}
      \toprule
                                                           &           & \multicolumn{8}{c}{image-\gls{auroc}}                                                                                                                                                             \\
      \cmidrule(l){3-10}
                                                           &           & \multicolumn{2}{c}{1-shot}            & \multicolumn{2}{c}{2-shot} & \multicolumn{2}{c}{4-shot} & \multicolumn{2}{c}{average}                                                                     \\
      \cmidrule(lr){3-4}
      \cmidrule(lr){5-6}
      \cmidrule(lr){7-8}
      \cmidrule(l){9-10}
                                                           & method    & {mean}                                & {std}                      & {mean}                     & {std}                       & {mean}         & {std}          & {mean}         & {std}          \\
      \midrule
      \multirow[c]{7}{*}{\rotatebox[origin=c]{90}{\mvt/}}  & PCore opt & 85.3                                  & 02.5                       & 88.9                       & \bfseries 00.5              & 91.9           & \bfseries 00.8 & 88.7           & \bfseries 01.3 \\
                                                           & PCore ori & 81.8                                  & \bfseries 01.1             & 86.4                       & 02.2                        & 89.5           & 01.7           & 85.9           & 01.7           \\
                                                           & WinCLIP   & \bfseries 93.1                        & \Uline{2.0}                & \bfseries 94.4             & \Uline{1.3}                 & \bfseries 95.2 & \Uline{1.3}    & \bfseries 94.2 & \Uline{1.5}    \\
                                                           & SPADE     & 81.0                                  & \Uline{2.0}                & 82.9                       & 02.6                        & 84.8           & 02.5           & 82.9           & 02.4           \\
                                                           & PaDiM     & 76.6                                  & 03.1                       & 78.9                       & 03.1                        & 80.4           & 02.5           & 78.6           & 02.9           \\
                                                           & GraphCore & \Uline{89.9}                          & {-}                        & \Uline{91.9}               & {-}                         & \Uline{92.9}   & {-}            & \Uline{91.6}   & {-}            \\
                                                           & RegAD     & 82.4                                  & {-}                        & 85.7                       & {-}                         & 88.2           & {-}            & 85.4           & {-}            \\
      \cmidrule{2-10}
      \multirow[c]{5}{*}{\rotatebox[origin=c]{90}{\visa/}} & PCore opt & \Uline{83.6}                          & \bfseries 01.6             & \bfseries 86.3             & \Uline{1.2}                 & \bfseries 88.8 & \Uline{0.7}    & \bfseries 86.4 & \bfseries 01.2 \\
                                                           & PCore ori & 76.7                                  & \Uline{2.4}                & 80.0                       & \bfseries 00.9              & 83.2           & \bfseries 00.6 & 80.0           & \Uline{1.3}    \\
                                                           & WinCLIP   & \bfseries 83.8                        & 04.0                       & \Uline{84.6}               & 02.4                        & \Uline{87.3}   & 01.8           & \Uline{85.2}   & 02.7           \\
                                                           & SPADE     & 79.5                                  & 04.0                       & 80.7                       & 05.0                        & 81.7           & 03.4           & 80.6           & 04.1           \\
                                                           & PaDiM     & 62.8                                  & 05.4                       & 67.4                       & 05.1                        & 72.8           & 02.9           & 67.7           & 04.5           \\
      \bottomrule
   \end{tabular}
\end{table}

\subsection{Limitations}
First, we did not optimize other \gls{ad}/\gls{as} methods besides PatchCore, but plan to do so in future work.
Due to compute constraints, we moreover did not ablate over the layer selection on the backbone.
However, doing so on a per-category basis was shown to improve \gls{ad} performance \cite{Heckler2023ExploringImportancePretrained}, and the optimal layers might furthermore depend on the number of shots k used.
Next, we did not reimplement all methods in our comparison with the state of the art, but rather took the values from the corresponding sources directly.
Unfortunately, most proposed methods do not provide reference implementations \cite{Jeong2023WinCLIPZero/few,Xie2023PushingLimitsFewshot}, and we found some to be vague in their description \cite{Xie2023PushingLimitsFewshot}, further increasing the difficulty of implementing them.
Nonetheless, this should be done for a fair comparison, and we plan to do so in future work.
Last, we did not perform a root-cause analysis on why anti-aliasing boosts \gls{ad}/\gls{as} performance in the few/many-shot settings.
We will therefore acquire a suitable dataset with higher variation in part positioning (i.e.\ translations, potentially also rotations) as a first step, rather than introducing these variations artificially (done e.g.\ in \cite{Artola2023GLADGlobalLocal}).
Afterwards, we will benchmark different constraints, i.e.\ translation and rotation-equivariant \glspl{cnn} \cite{Zhang2019Makingconvolutionalnetworks,Weiler2018LearningSteerableFilters,Cesa2022ProgramBuildEN,Michaeli2023AliasFreeConvnets}, as well as their \acrlong{vit} counterparts \cite{Qian2021BlendingAntiAliasing,Xu2023$E2$EquivariantVision}, on said dataset.

\section{Conclusion}
We performed a comprehensive study to optimize the various hyperparameters of PatchCore for the few/many-shot \gls{ad}/\gls{as} settings.
The study revealed that the feature extractor and the input resolution are the most important hyperparameters to tune.
Additionally, we showed that image augmentations can be helpful, but need to be carefully designed using domain knowledge or one risks actually decreasing performance.
Next, we showed that coreset subsampling combined with an adequate augmentation scheme allows one to keep inference times low while maintaining the benefits of augmenting the dataset.
Combining all of this, we were able to achieve state-of-the-art \gls{ad} performance on the challenging \visa/ dataset in the few-shot \gls{ad} setting, demonstrating the merit of adapting pre-existing \gls{ad}/\gls{as} methods to the few/many-shot settings.
Finally, we showed that leveraging feature extractors with a stronger inductive bias such as anti-aliasing improves performance in the few/many-shot \gls{ad}/\gls{as} settings, revealing a promising avenue for future research.

   {\small
      \bibliographystyle{ieee_fullname}
      \bibliography{./literatur/lit}
   }

\clearpage

\appendix

\section{Appendix}

\subsection{Defect Size}\label{appendix_defect_size}
\begin{figure}
   \centering
   \begin{subfigure}[b]{1.0\linewidth}
      \centering
      \includegraphics[width=1.0\linewidth]{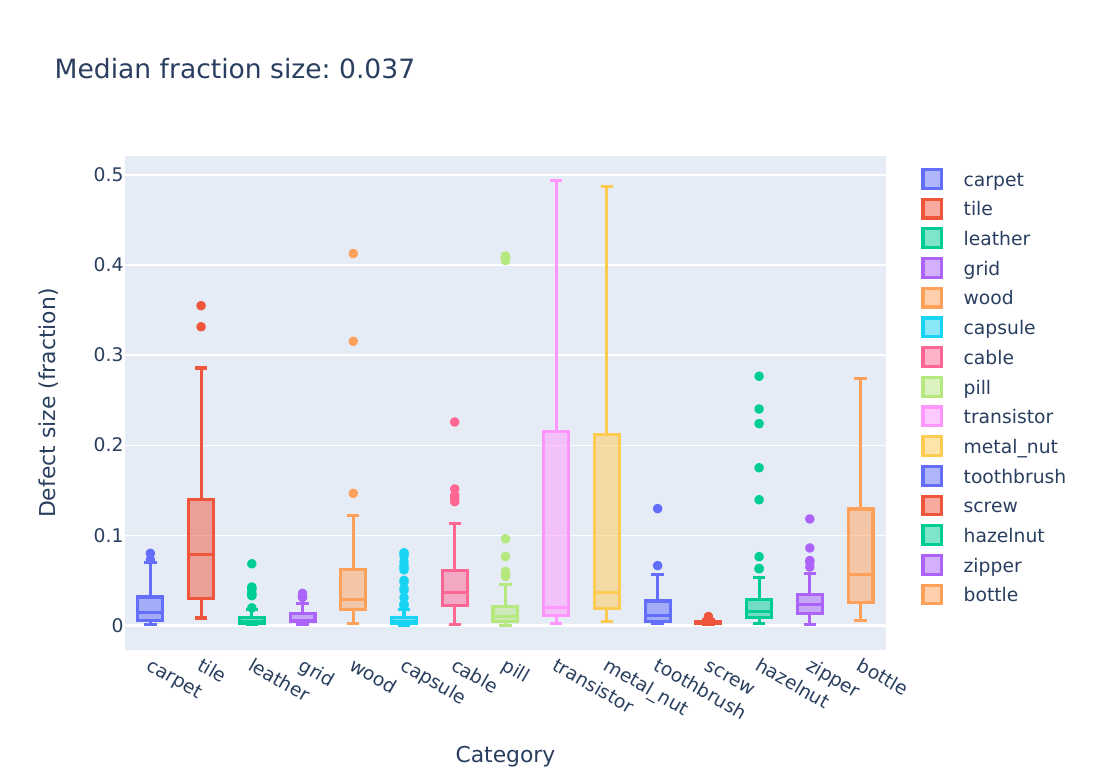}
      \caption{}
      \label{fig:defect_size_mvtec}
   \end{subfigure}
   \hfill
   \begin{subfigure}[b]{1.0\linewidth}
      \centering
      \includegraphics[width=1.0\linewidth]{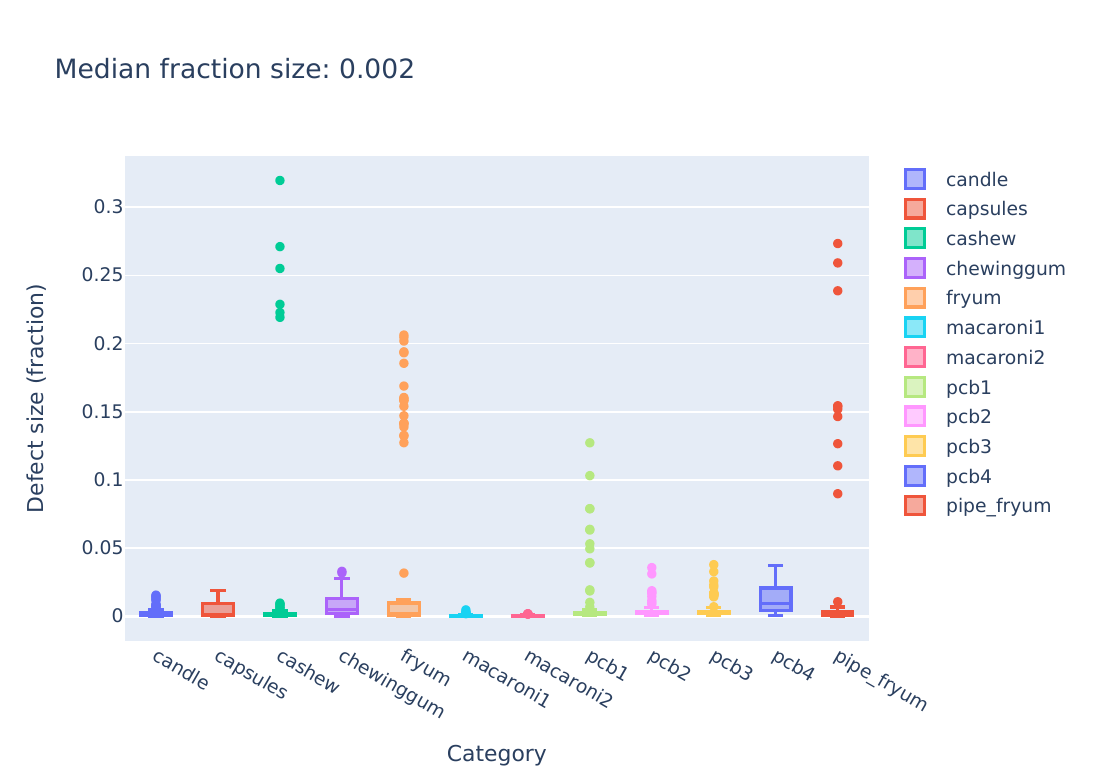}
      \caption{}
      \label{fig:defect_size_visa}
   \end{subfigure}
   \caption{Distribution of defect sizes for (a) \mvt/ and (b) \visa/.}
\end{figure}

\Cref{fig:defect_size_mvtec,fig:defect_size_visa} show the distribution of defect size as a fraction of the image's pixels, for both \mvt/ and \visa/ categories.
One can see from these results that defects in \visa/ are generally much smaller than in \mvt/, which can partially explain its increased difficulty.

\subsection{Influence of Dataset, Feature Extractor and Input Resolution}

\subsubsection{Influence of Dataset and Input Resolution}\label{appendix_results_scale-vs-datset}
\Cref{table:scale_dataset_comparison} shows the AUHPROC on both \mvt/ and \visa/.
Looking more closely at this performance, we can see that the trend of increase in performance with an increase in scale is stronger in \visa/ compared to \mvt/.
One may argue that this is due to the fact that \visa/ contains smaller defects than \mvt/, and that higher resolution might thus better facilitate their detection (see also \cref{appendix_defect_size}).

\begin{table*}[t]
   \centering
   \caption{Performance between different input scales on \mvt/ and \visa/ datasets. For each dataset, we report the mean of category-wise AUHPROC after averaging over feature extractors and seeds/runs. We also report the relative performance gain/loss between input scales 0.5x, 1.5x and 2.0x compared to the original input scale 1x.}
   \label{table:scale_dataset_comparison}
   \begin{tabular}{lrrrrrrr}
      \toprule
                                      & \multicolumn{7}{c}{AUHPROC}                                                                                                                     \\
      \cmidrule(lr){2-8}
      \multicolumn{1}{l}{input scale} & \multicolumn{1}{c}{1.0x (original scale)} & \multicolumn{2}{c}{0.5x} & \multicolumn{2}{c}{1.5x} & \multicolumn{2}{c}{2.0x}                      \\
      \cmidrule(lr){2-2}
      \cmidrule(lr){3-4}
      \cmidrule(lr){5-6}
      \cmidrule(lr){7-8}
                                      & mean                                      & mean                     & diff                     & mean                     & diff & mean & diff \\
      \midrule
      \mvt/                           & 63.3                                      & 52.1                     & -11.2                    & 66.9                     & 3.6  & 68.3 & 5.0  \\
      \visa/                          & 45.5                                      & 34.6                     & -10.9                    & 50.2                     & 4.7  & 53.0 & 7.5  \\
      \bottomrule
   \end{tabular}
\end{table*}

\subsubsection{Influence of Feature Extractor and Dataset}\label{appendix_results_featureextractor-vs-datset}
\begin{figure}
   \centering
   \begin{subfigure}[b]{1.0\linewidth}
      \centering
      \includegraphics[width=1.0\linewidth]{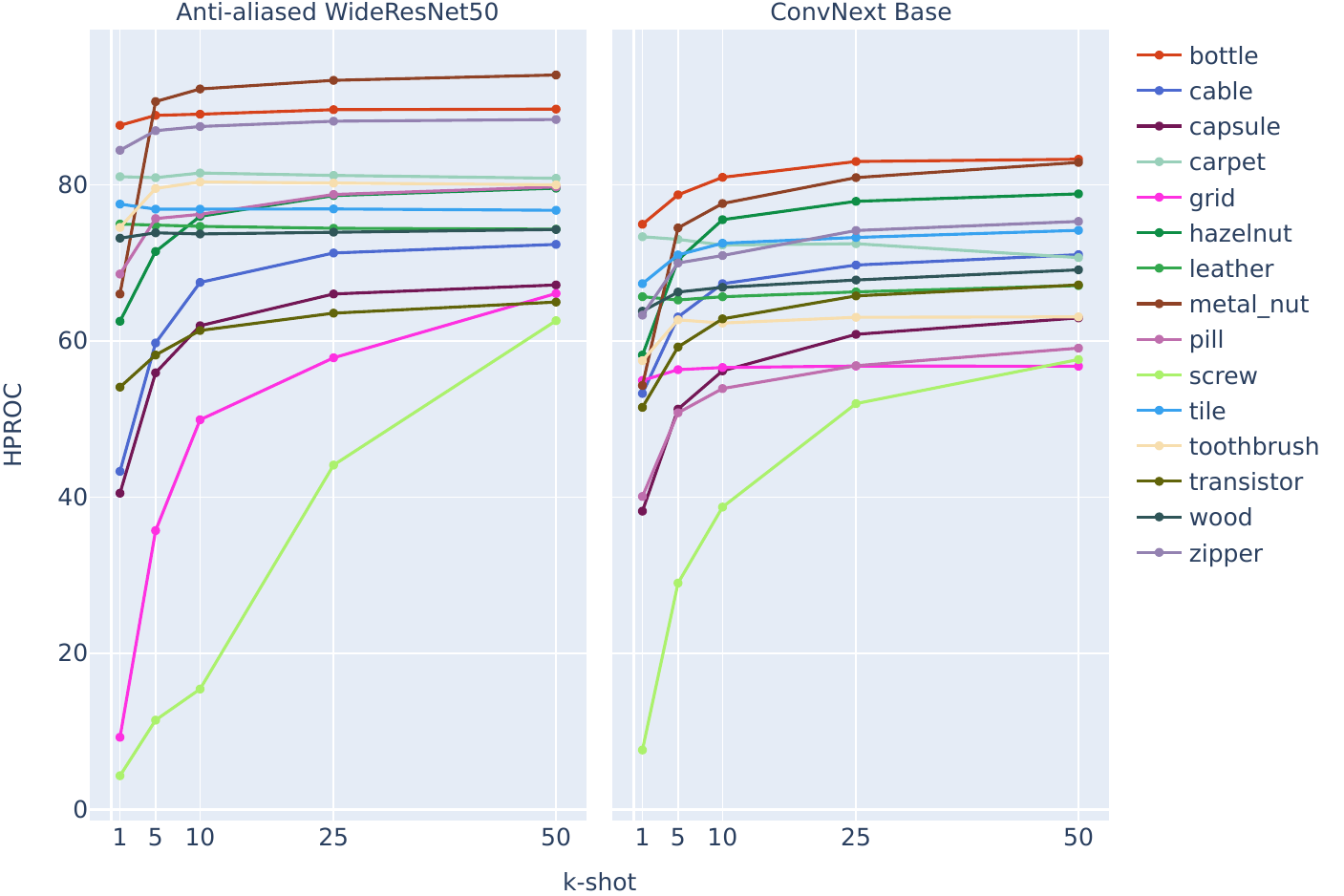}
      \caption{}
      \label{fig:best_fx_mvtec}
   \end{subfigure}
   \hfill
   \begin{subfigure}[b]{1.0\linewidth}
      \centering
      \includegraphics[width=1.0\linewidth]{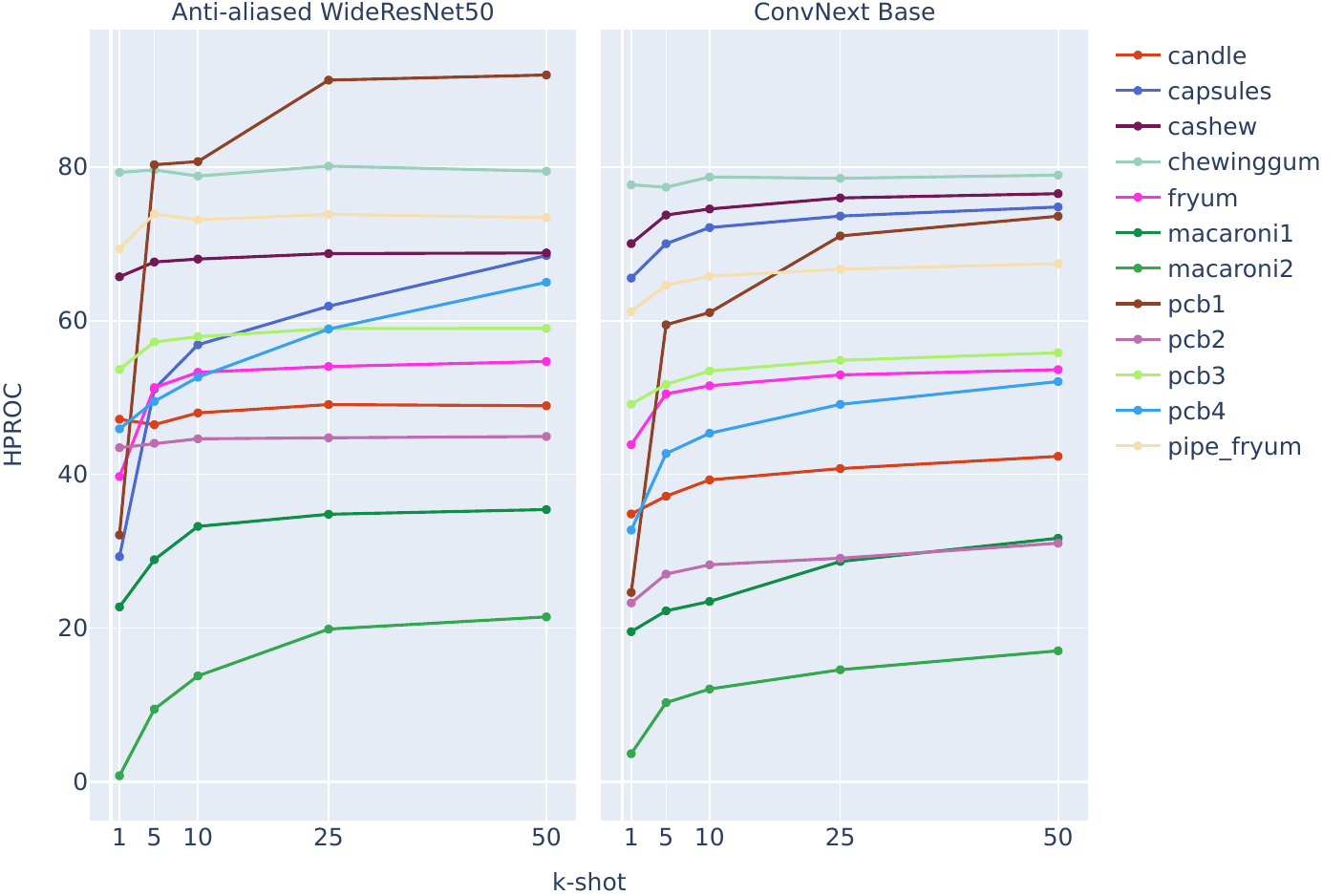}
      \caption{}
      \label{fig:best_fx_visa}
   \end{subfigure}
   \caption{HPROC across k-shots for (a) \mvt/ and (b) \visa/ dataset, using Anti-aliased WideResNet50 (left) and ConvNext Base (right) as feature extractors at 2x their original input scale.}
\end{figure}

We plot category-wise HPROC vs.\ k-shot curves for two different feature extractors (Anti-aliased WideResNet50 and ConvNext Base), at their best-performing scale (2x), in \cref{fig:best_fx_mvtec,fig:best_fx_visa}.
We see that even though the Anti-aliased WideResNet50 performs best overall, it does not perform best in all categories and shots (e.g.\ "capsules" and "cashew" of \visa/).

\subsubsection{Influence of Feature Extractor and Scale}\label{appendix_results_featureextractor-vs-scale}
\begin{figure}[t]
   \centering
   \includegraphics[width=1.0\linewidth]{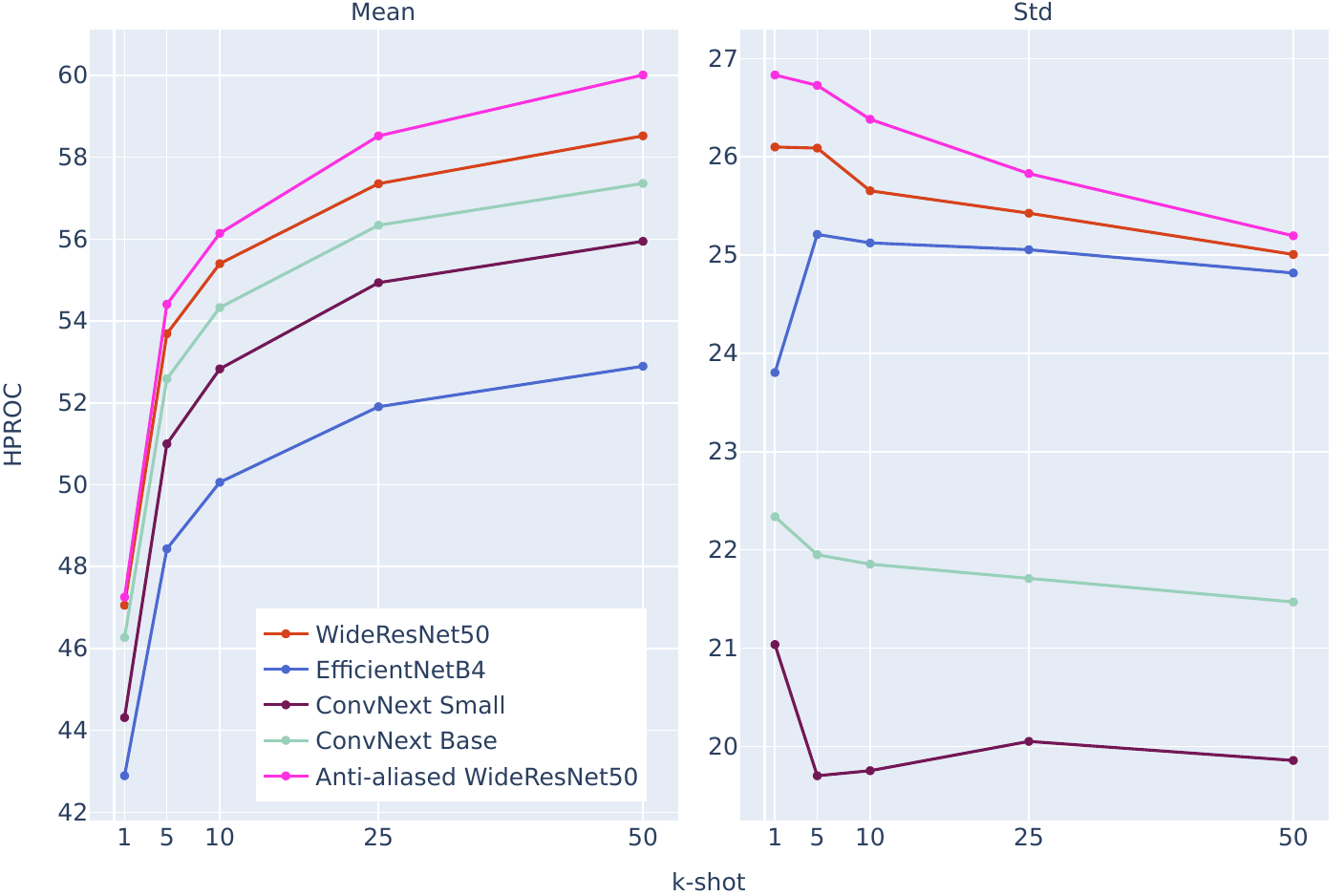}
   \caption{Mean and standard deviation of HPROC across k-shots for different feature extractors, averaged over datasets, input scales, and seeds/runs.}
   \label{fig:feature_extractors}
\end{figure}

We show in \cref{fig:feature_extractors} the HPROC vs. k-shot curves for each backbone.
Specifically, we compute the mean and standard deviation of HPROC across categories, scales and seeds/runs.

\subsection{Data Augmentation}\label{appendix_influence_aug}
We further investigate the effect of augmentations on \mvt/ without using the Sharpen augmentation and only keep Affine, BrightnessContrast and Blur augmentations.
We employ an Anti-aliased WideResNet50 at 2x its original scale.
We can see in \cref{table:mvtec_no_sharpen_augmentation_comparison} that applying the three augmentations allows us to surpass the case where no augmentations are applied, and that disabling any one of the augmentations greatly decreases performance.
This indicates that the combination of these three augmentations is important to actually increase performance.

\begin{table}[t]
   \centering
   \caption{Performance between different augmentation settings on \mvt/ dataset.
      We disabled each respective augmentation while keeping the remaining ones active.
      We report the mean of category-wise AUHPROC generated after averaging over seeds/runs.
      We also report the relative gain/loss of different augmentation combinations compared to disabling augmentation.
      Abbreviations:
      \# augs~=~number of augmentations,
      A~=~Affine,
      BC~=~BrightnessContrast,
      B~=~Blur.
   }

   \begin{tabular}{lllllrr}
      \toprule
                            & \multicolumn{3}{c}{augmentation type} & \multicolumn{2}{c}{AUHPROC}                        \\
      \cmidrule(lr){2-4}
      \cmidrule(lr){5-6}
      \# augs               & A                                     & BC                          & B      & mean & diff \\
      \midrule
      0                     & \xmark                                & \xmark                      & \xmark & 67.6 & 0.0  \\
      \cmidrule(lr){1-6}
      \multirow[t]{4}{*}{8} & \xmark                                & \cmark                      & \cmark & 67.3 & -0.3 \\

                            & \multirow[t]{3}{*}{\cmark}            & \xmark                      & \cmark & 67.4 & -0.2 \\

                            &                                       & \multirow[t]{2}{*}{\cmark}  & \xmark & 67.2 & -0.4 \\
                            &                                       &                             & \cmark & 68.2 & 0.6  \\
      \bottomrule
   \end{tabular}
   \label{table:mvtec_no_sharpen_augmentation_comparison}
\end{table}

\subsection{Failure Case Analysis}\label{subsec:per_category_results}
\begin{figure*}
   \centering
   \begin{tabular}{llllll}
      cashew                                                        &
      \includegraphics[width=.15\linewidth]{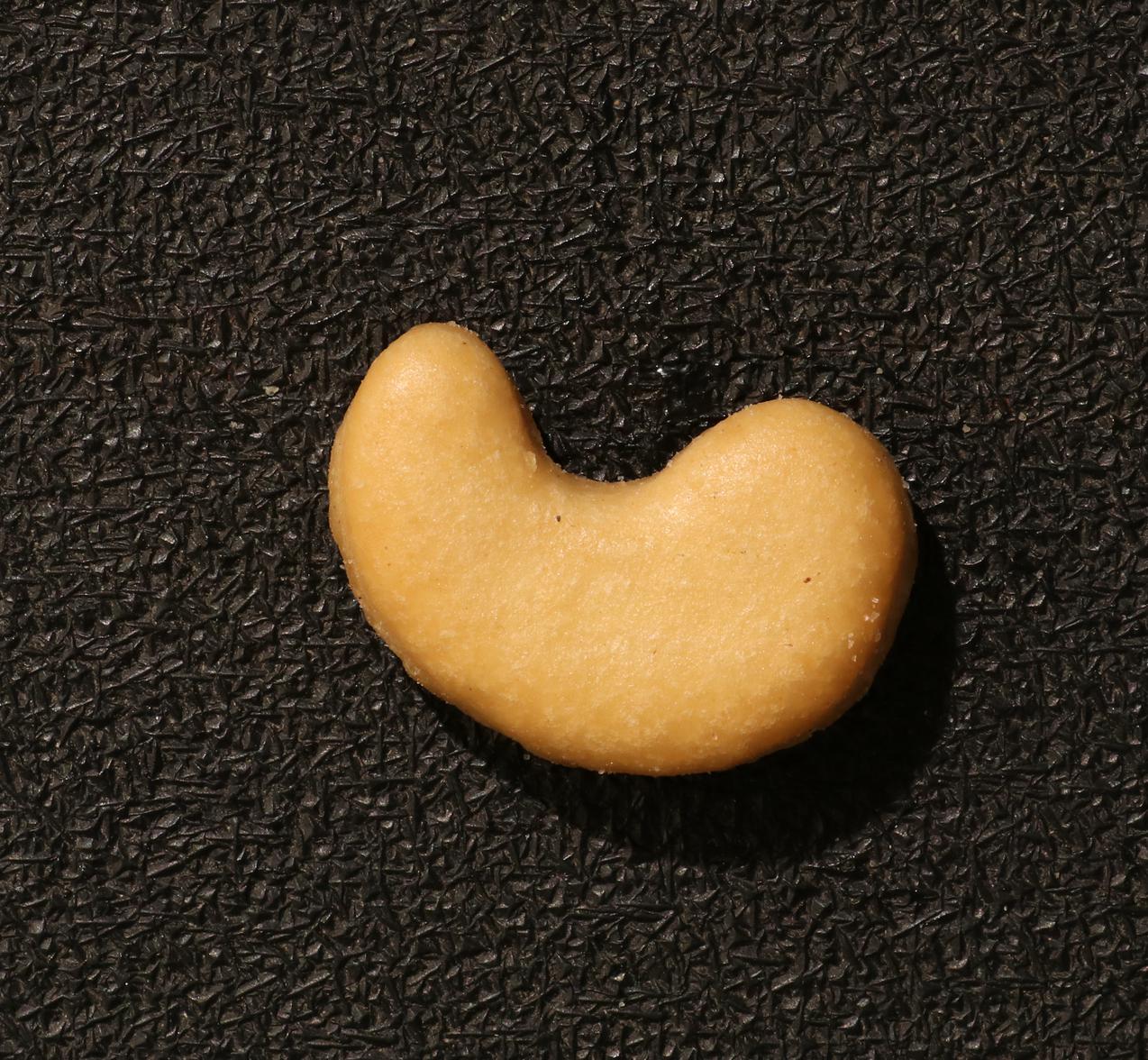}     & \includegraphics[width=.15\linewidth]{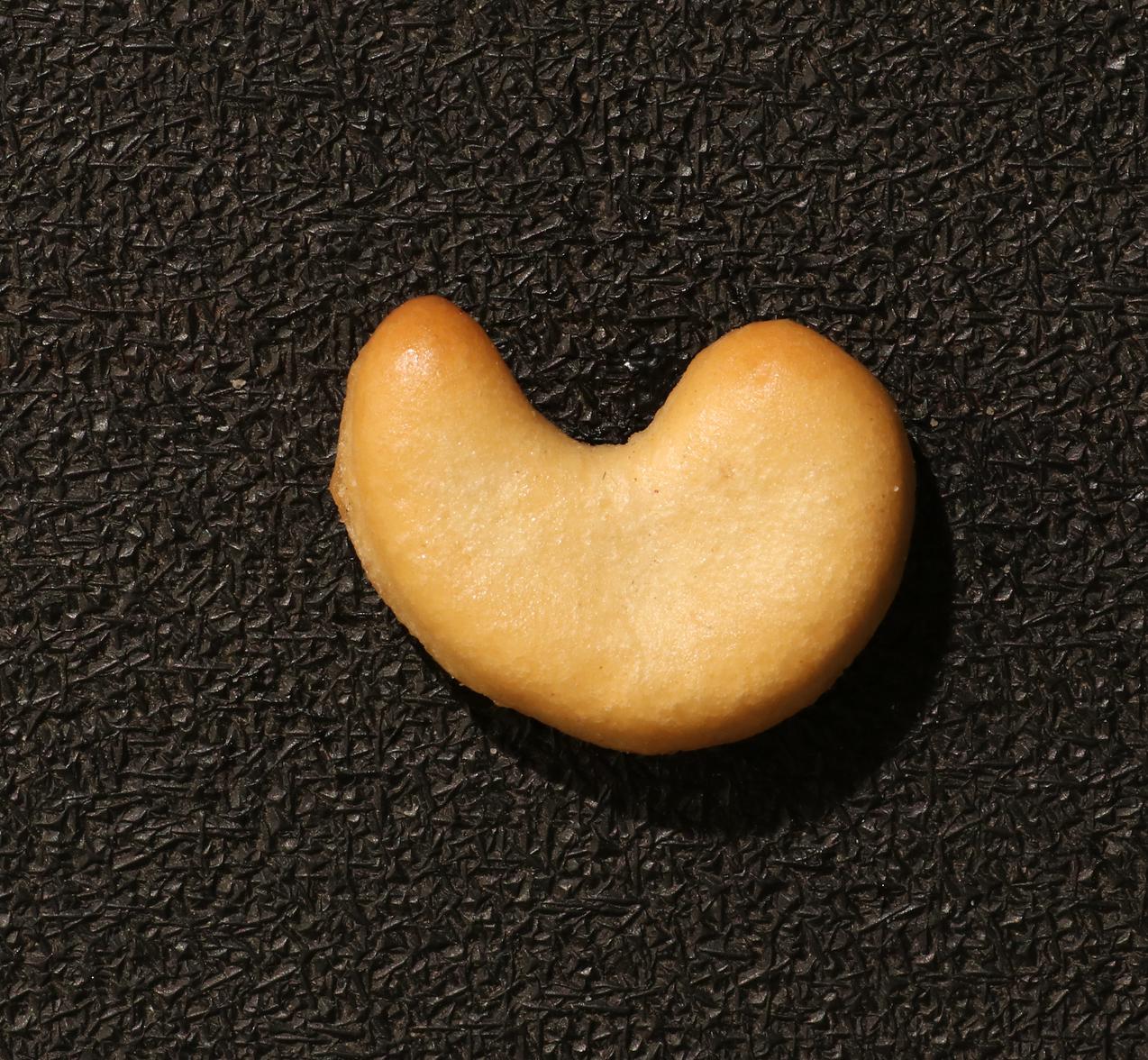}     & \includegraphics[width=.15\linewidth]{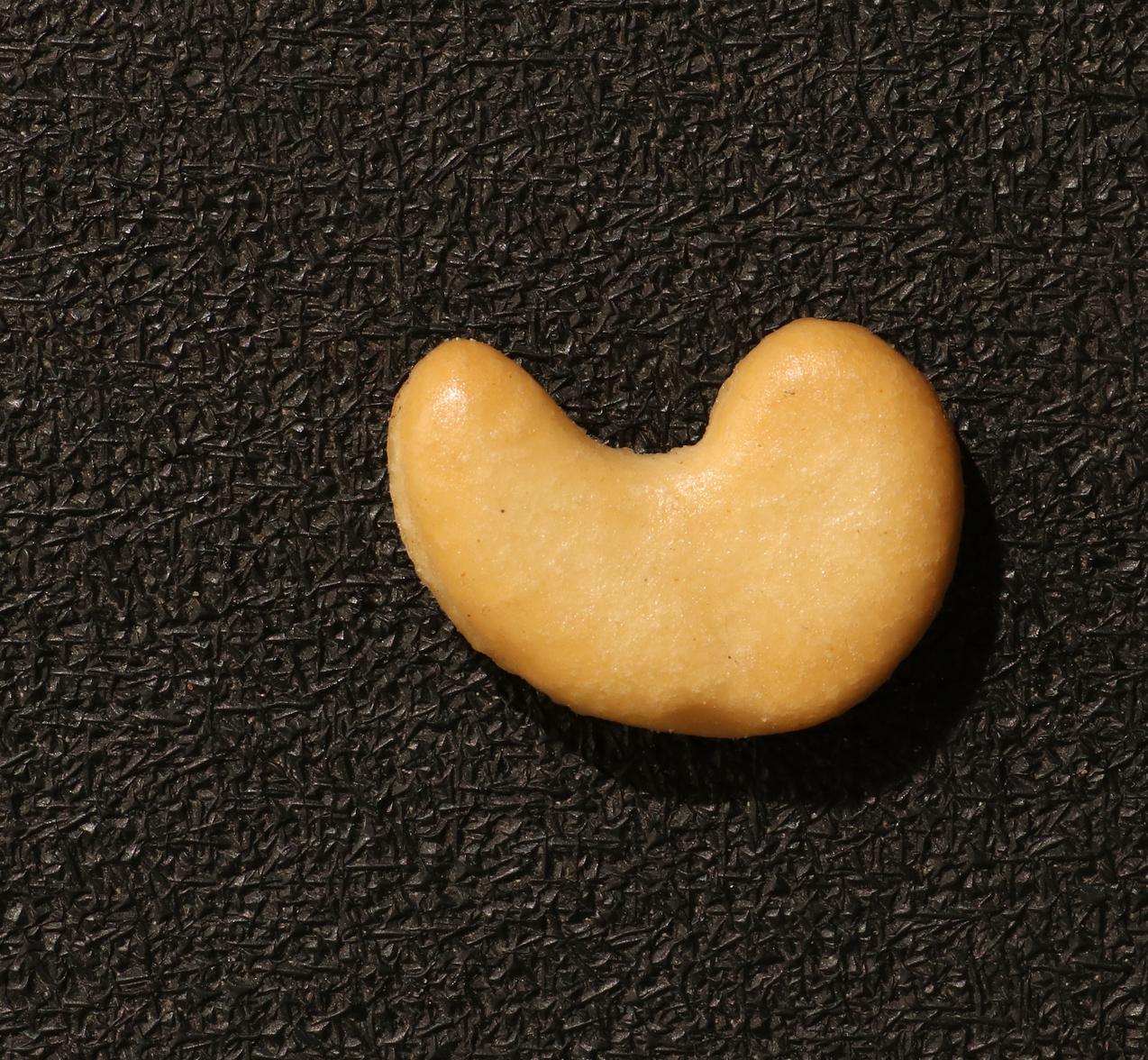} &
      \includegraphics[width=.15\linewidth]{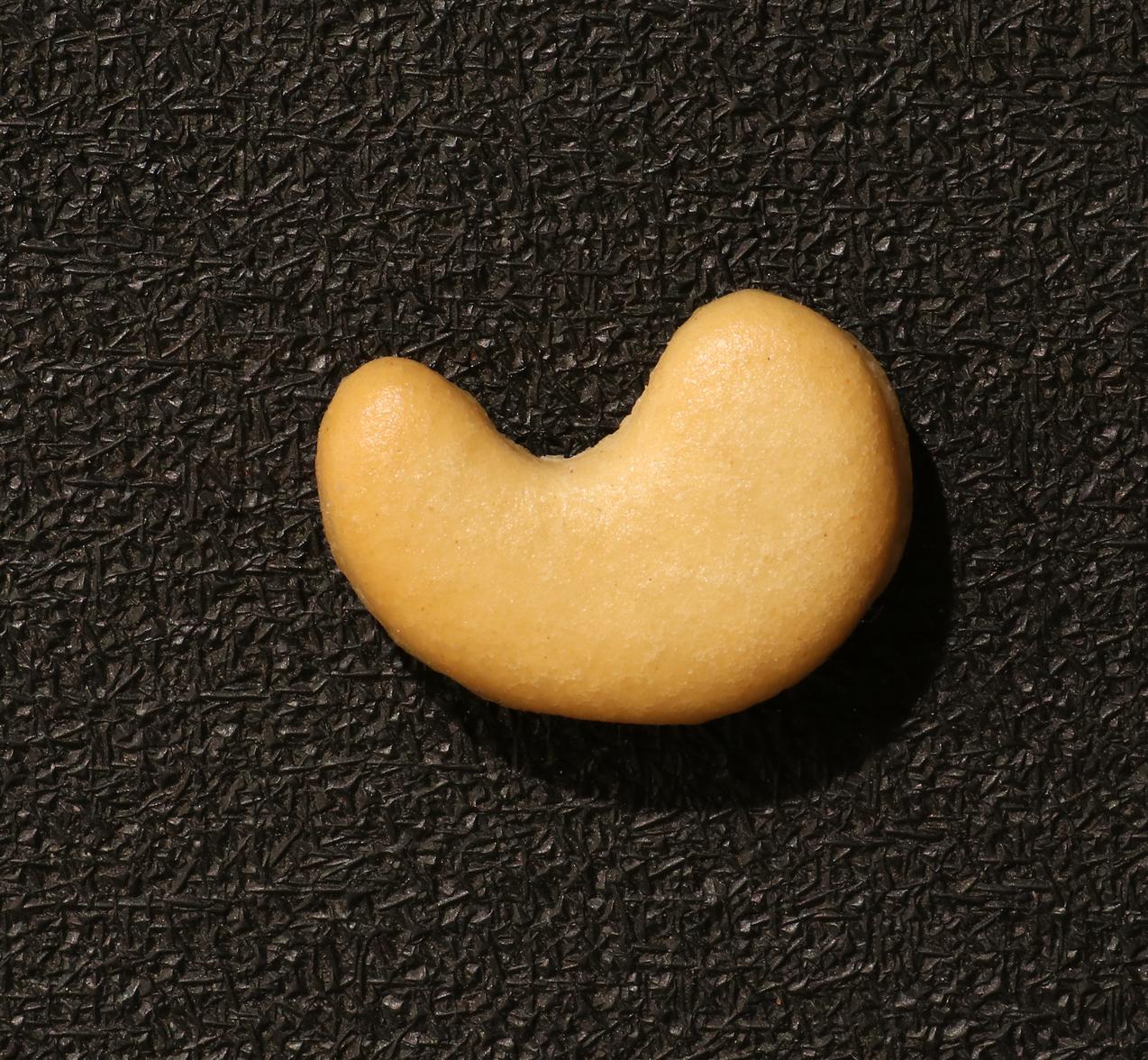}     &
      \includegraphics[width=.15\linewidth]{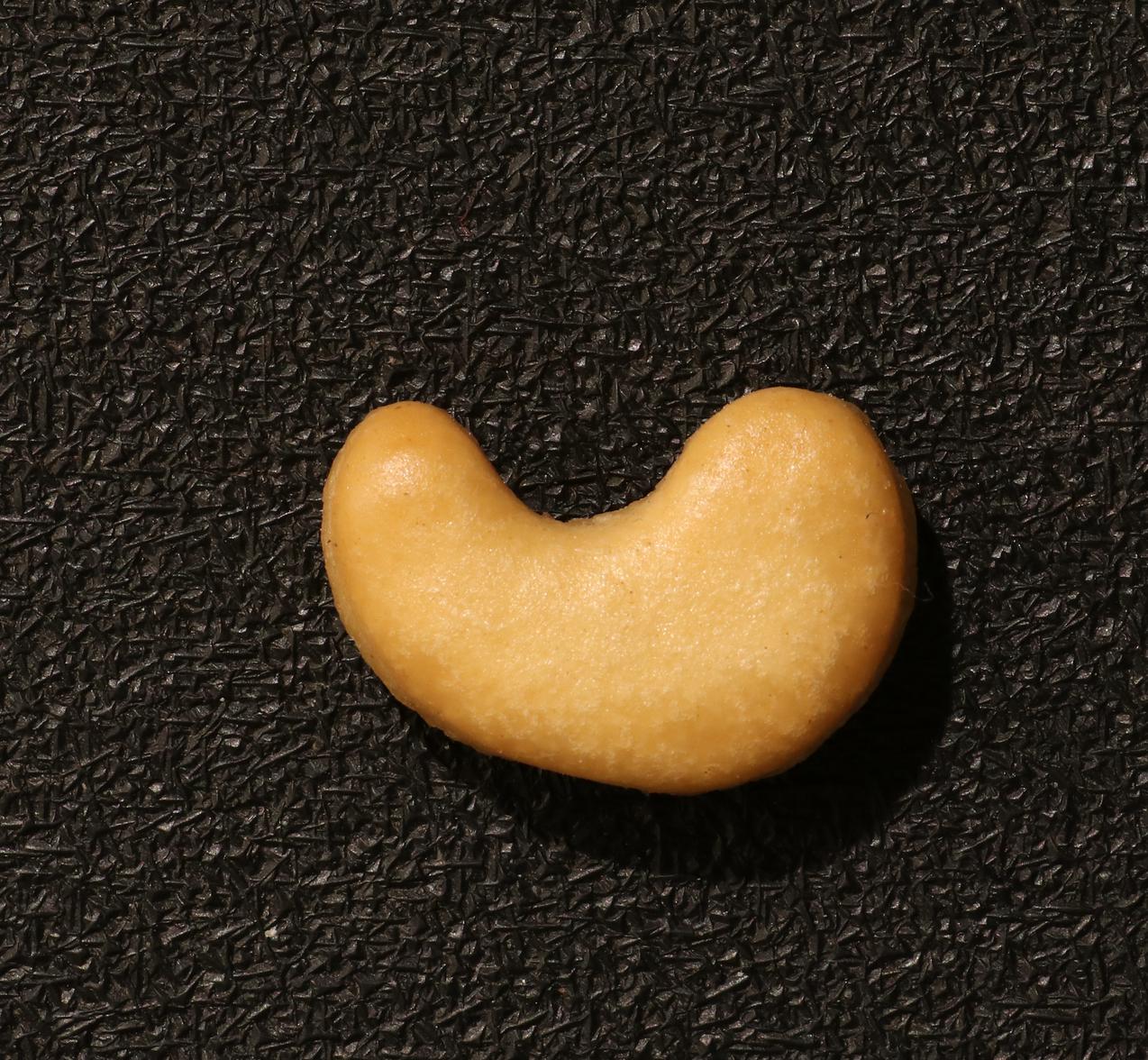}
      \\
      macaroni2                                                     &
      \includegraphics[width=.15\linewidth]{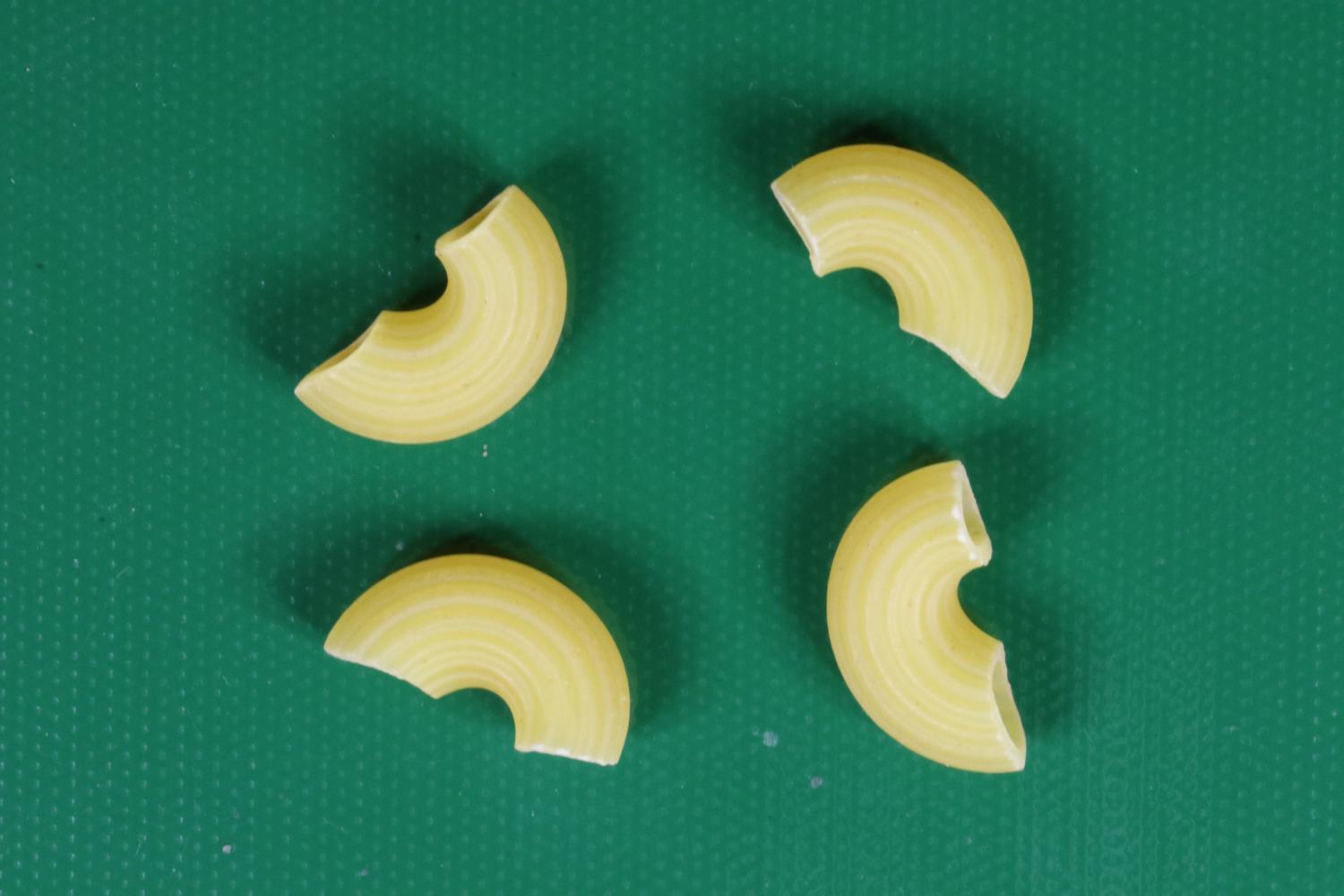} & \includegraphics[width=.15\linewidth]{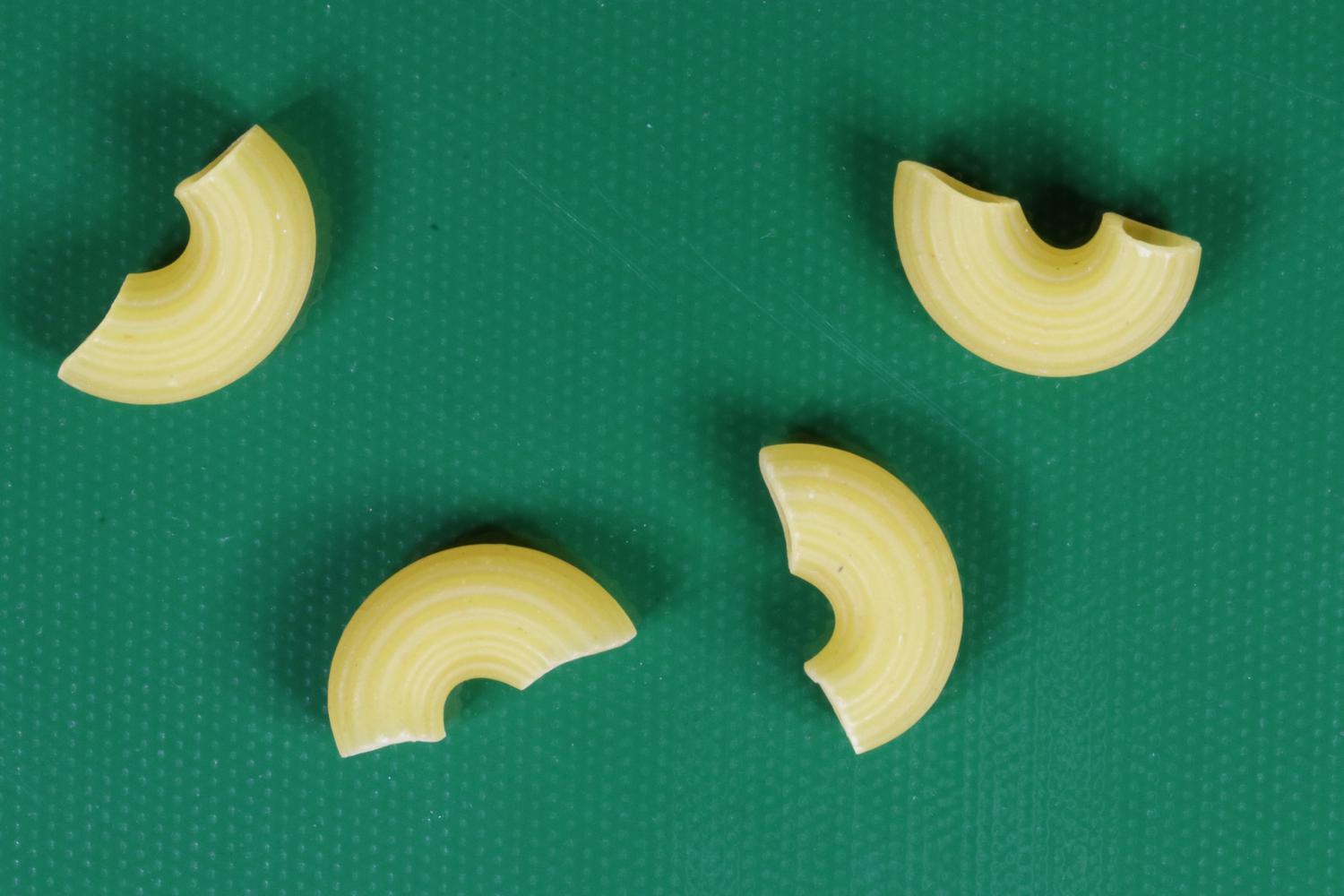} &
      \includegraphics[width=.15\linewidth]{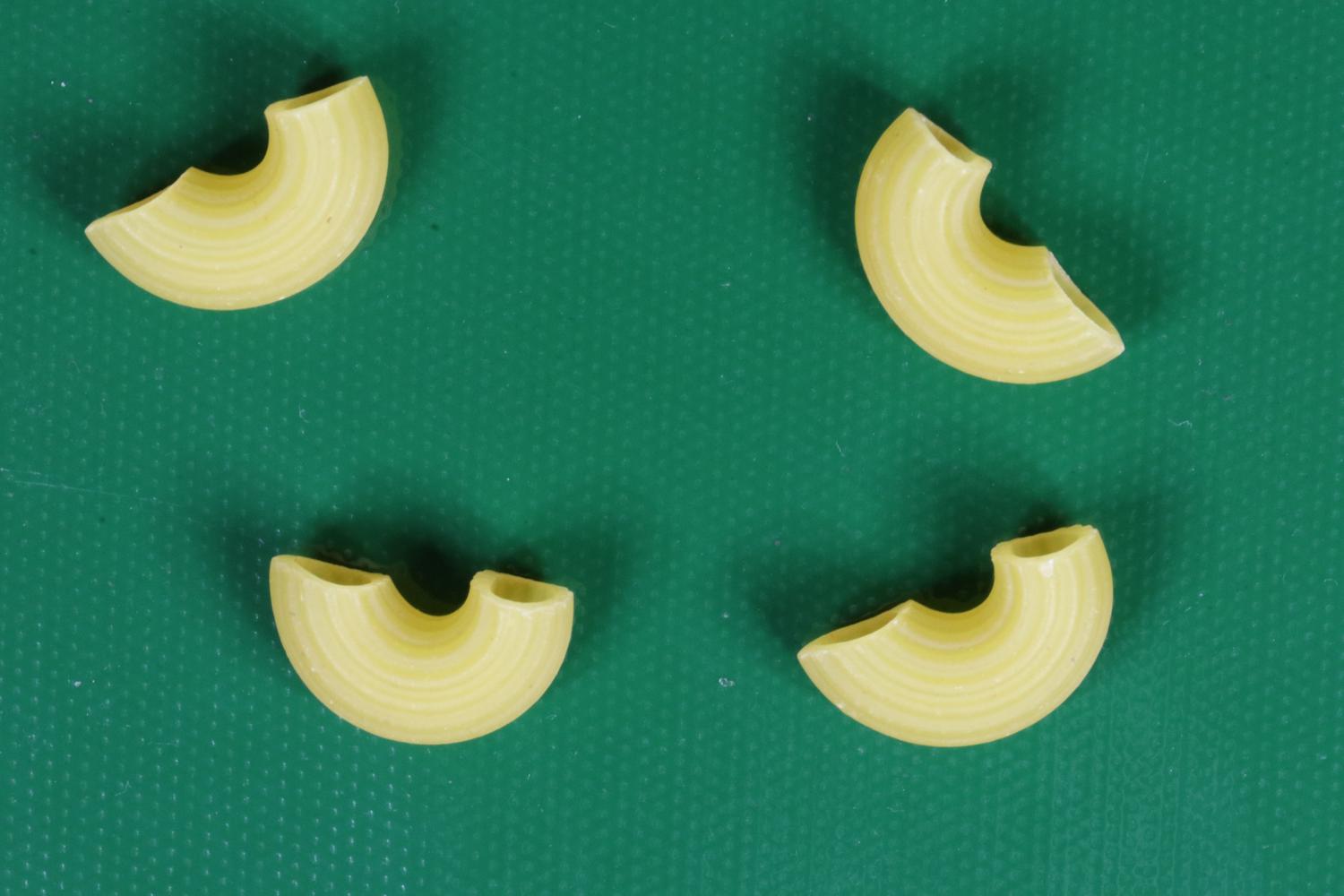} &
      \includegraphics[width=.15\linewidth]{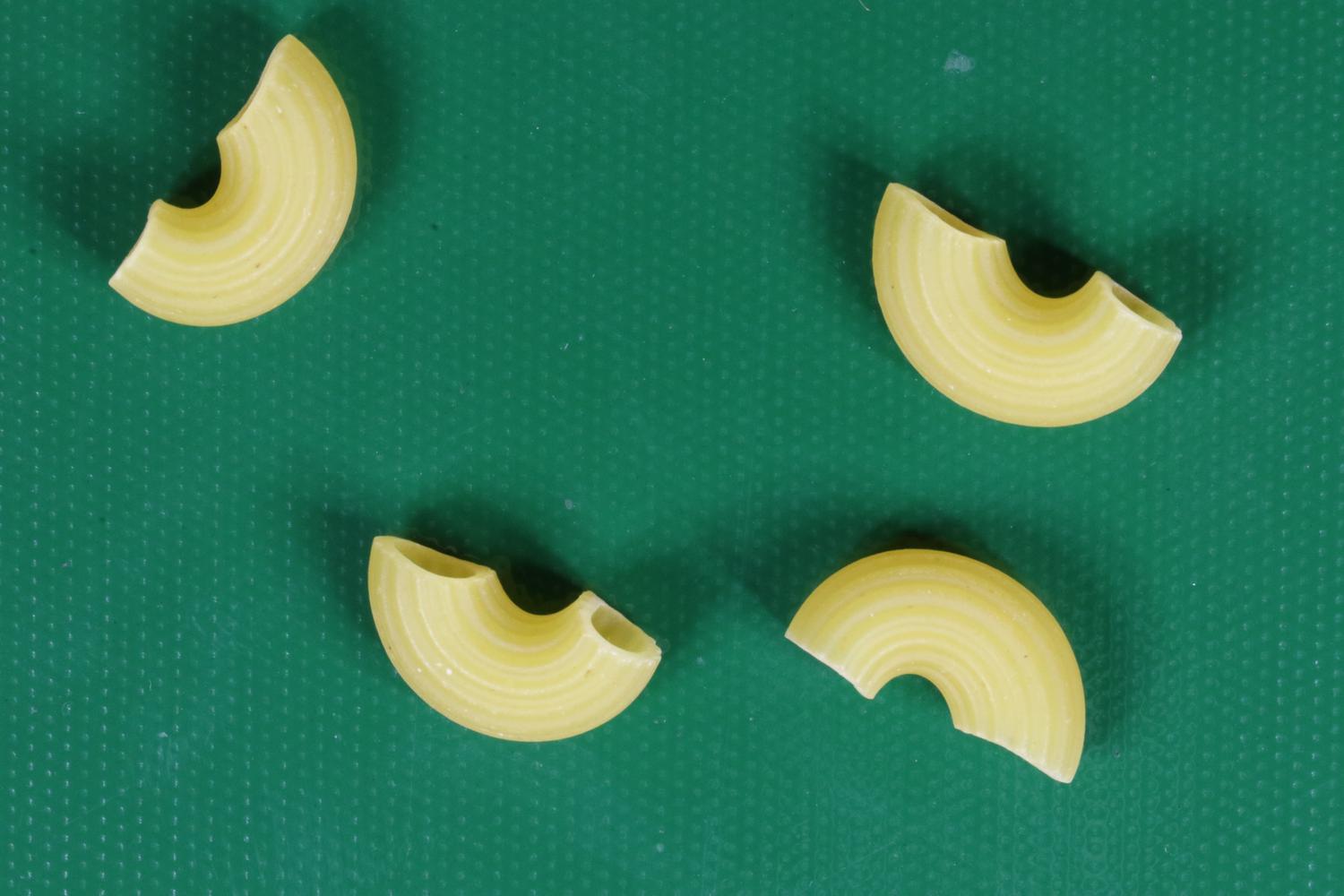} &
      \includegraphics[width=.15\linewidth]{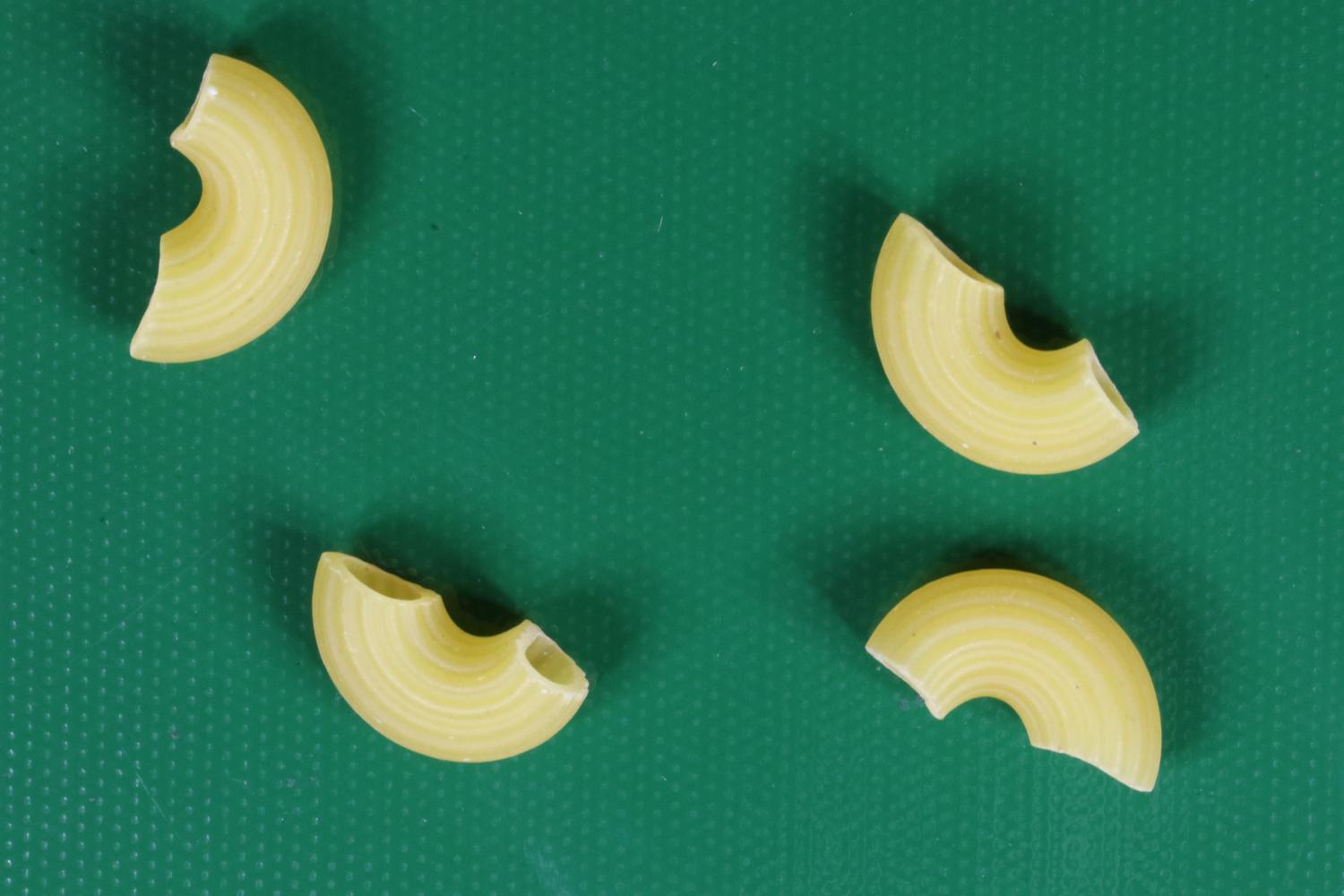}
      \\
      screw                                                         &
      \includegraphics[width=.15\linewidth]{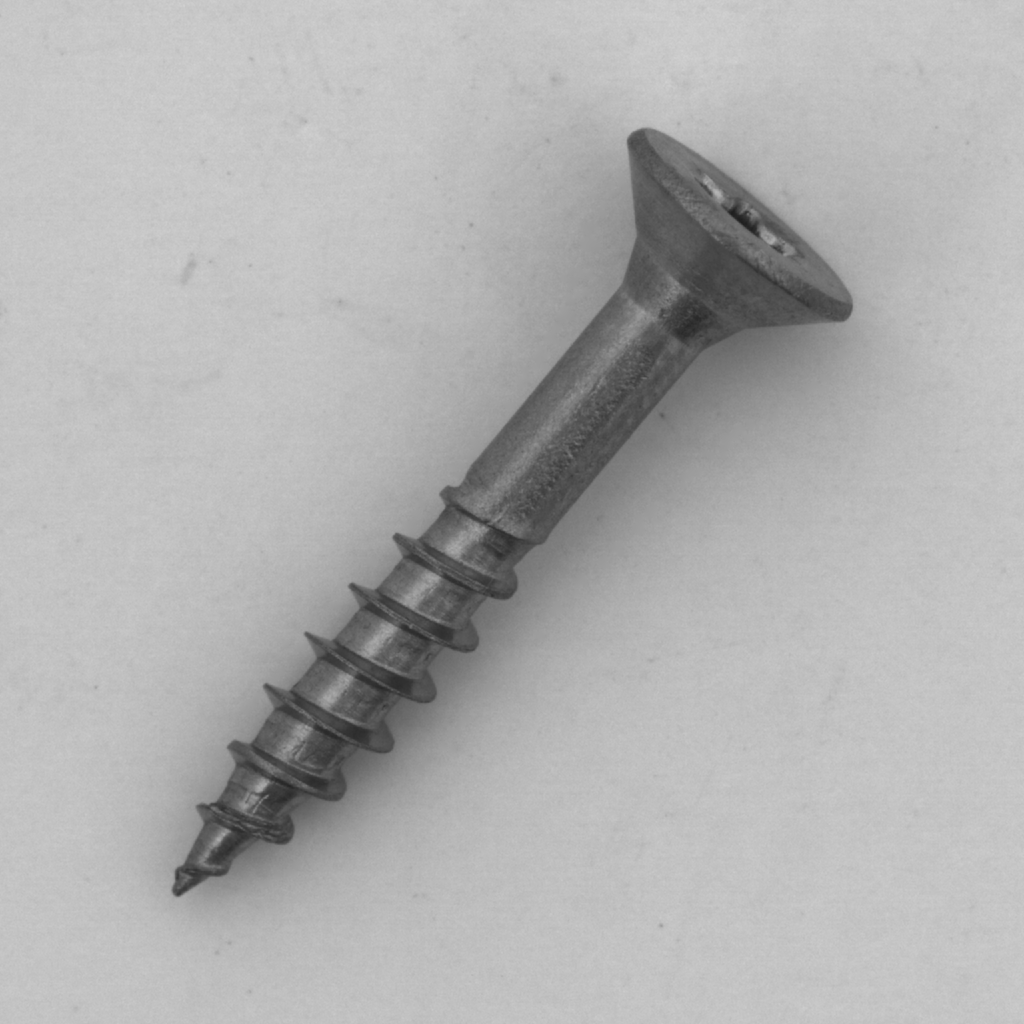}      &
      \includegraphics[width=.15\linewidth]{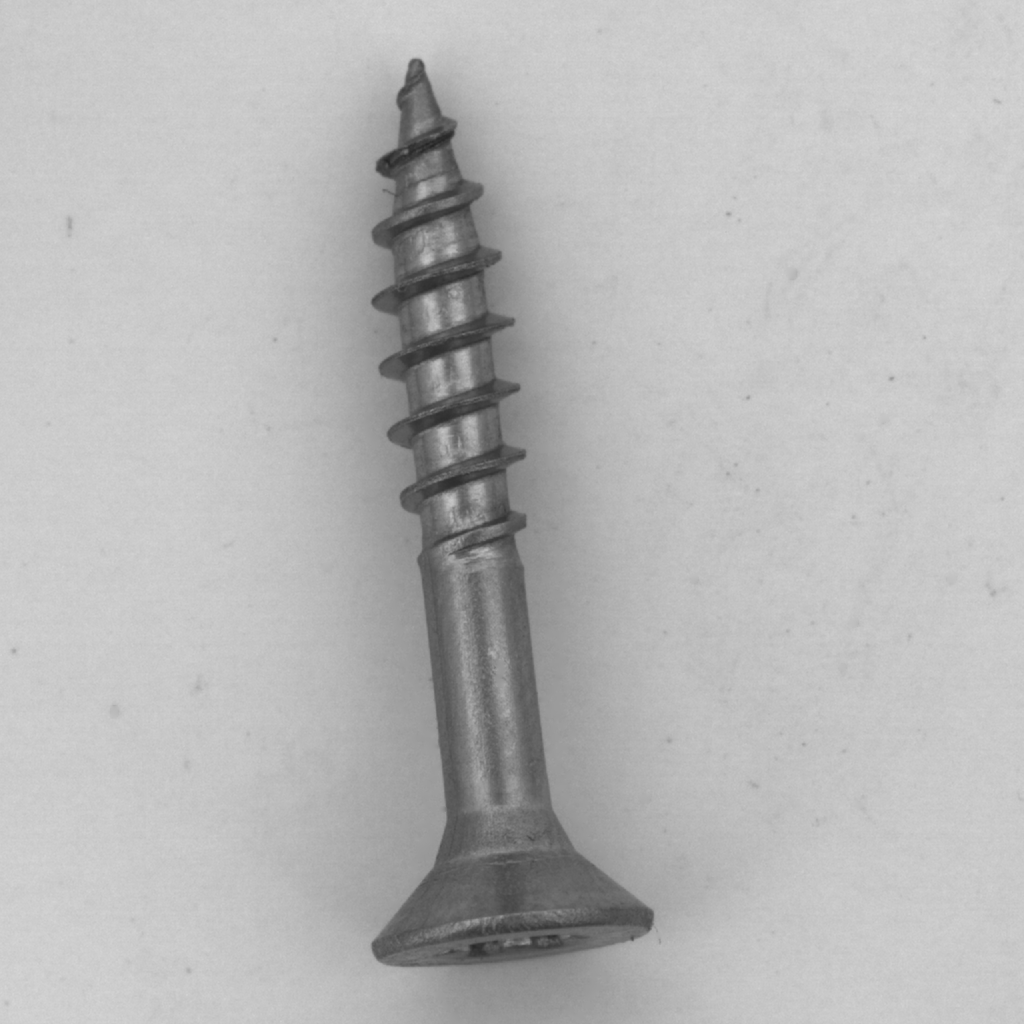}      &
      \includegraphics[width=.15\linewidth]{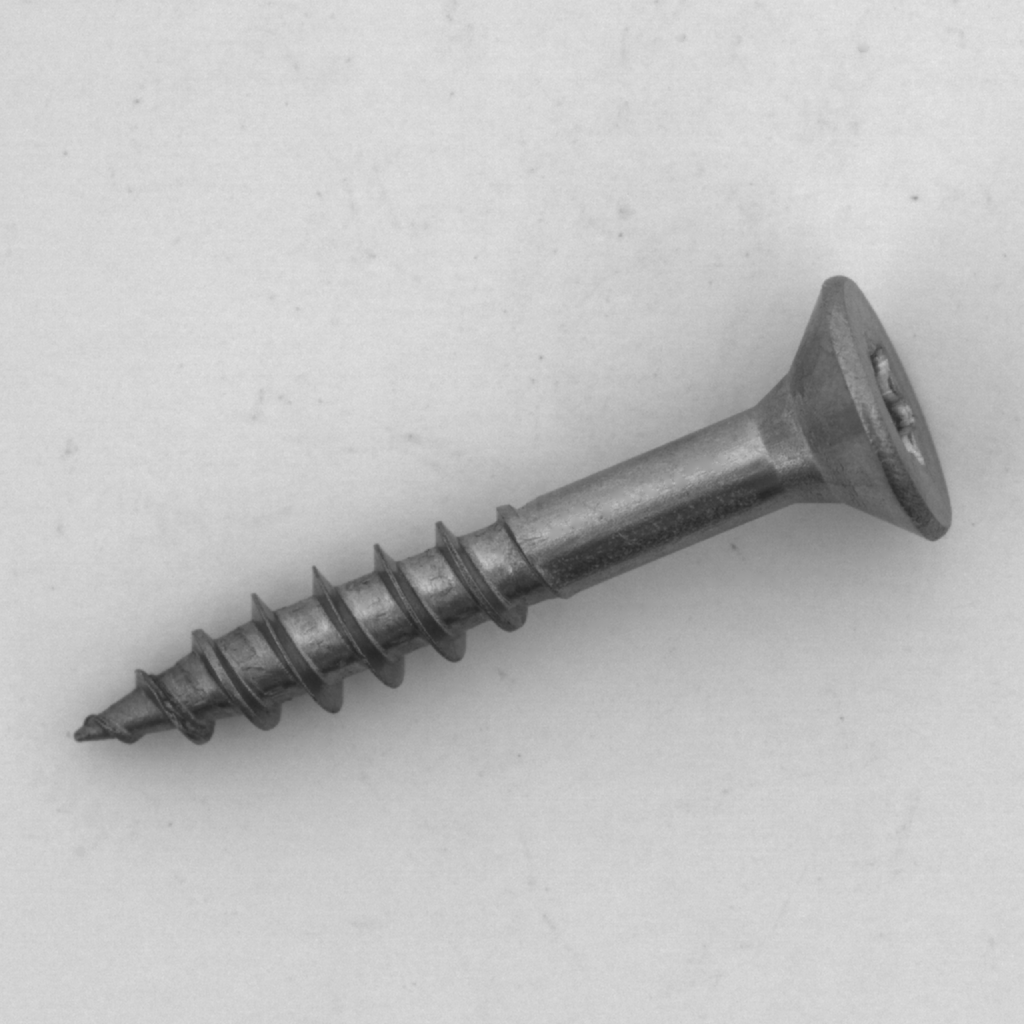}      &
      \includegraphics[width=.15\linewidth]{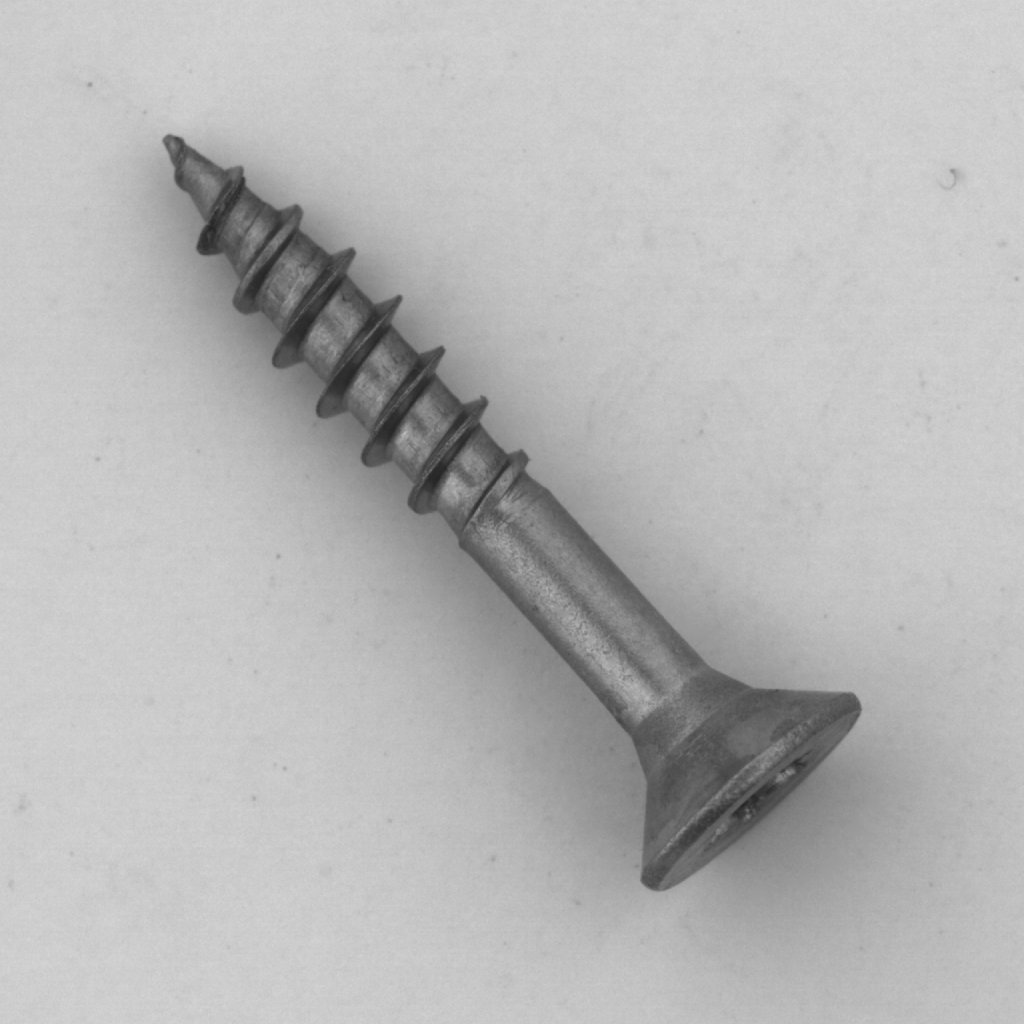}      &
      \includegraphics[width=.15\linewidth]{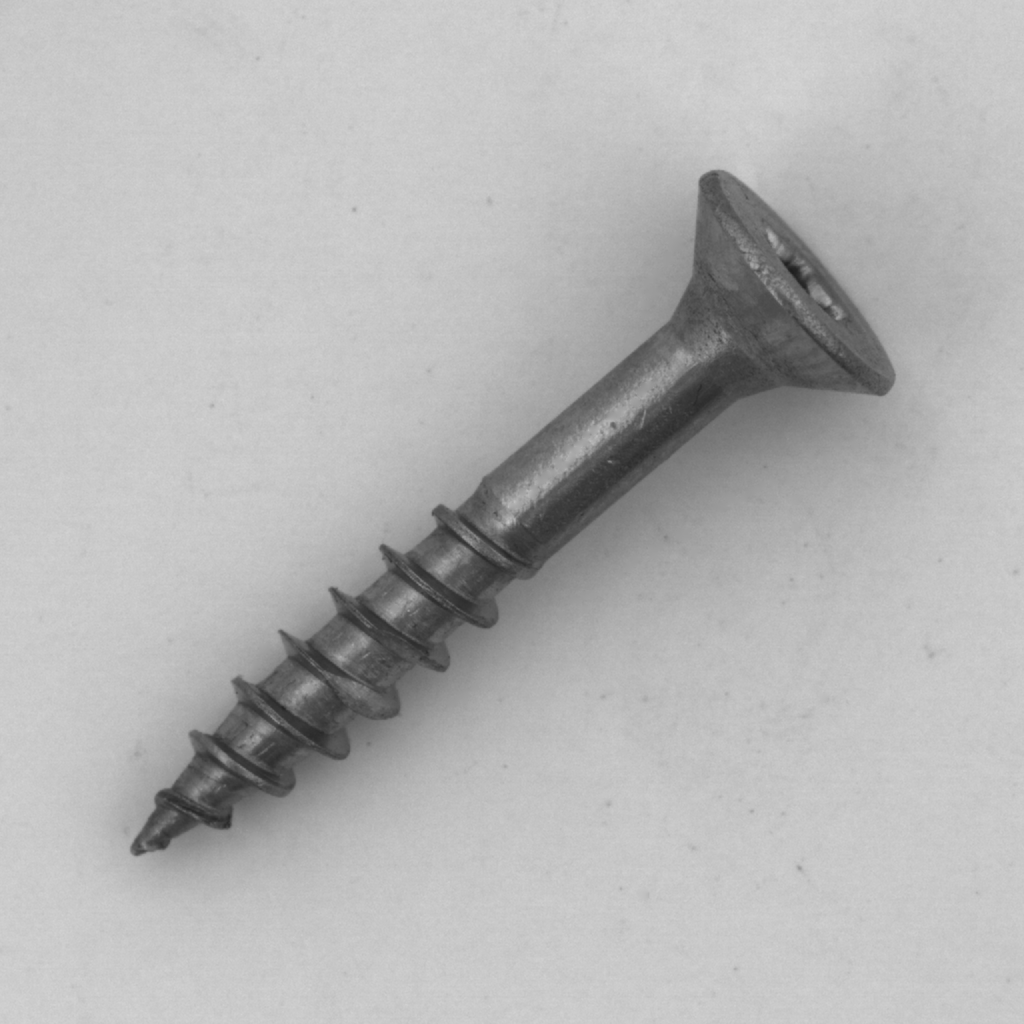}
   \end{tabular}
   \caption{Low degree of rotation for "cashew" compared to high degree of rotation for "macaroni2" and "screw".}
   \label{fig:rotation_degree}
\end{figure*}

We report category-wise image-\gls{auroc} scores of optimized PatchCore (see \cref{sec:sota}) for both \mvt/ and \visa/ across different k-shots to identify failure cases.
Assessing \cref{table:mvtec_per_category_results}, we see that certain categories where normal data contains strong rotations yield sub-optimal results (e.g.\ "screw" for \mvt/ and "macaroni2" on \visa/, see \cref{fig:rotation_degree} for some qualitative examples).
It is worth noting here that \glspl{cnn} are not rotation-equivariant by design.
This can be attributed to the inherent properties of the convolution operation, which relies on fixed filters to extract local features.
As a result, \glspl{cnn} struggle to generalize across different rotation angles, leading to a decreased performance.
Therefore, one should additionally investigate rotation-equivariant networks \cite{Weiler2018LearningSteerableFilters,Cesa2022ProgramBuildEN} as pre-trained feature extractor in future work for further performance improvements.

\begin{table*}[t]
   \centering
   \caption{Category-wise image-\gls{auroc} scores of optimized PatchCore for \mvt/ and \visa/ across different k-shots.
      We report the mean and standard deviation over 5 random seeds for each category.}
   \label{table:mvtec_per_category_results}
   \begin{tabular}{llrrrrrrrrrr}
      \toprule
                                                            &            & \multicolumn{10}{c}{image-\gls{auroc}}                                                                                                                                                         \\
      \cmidrule(l){3-12}
                                                            &            & \multicolumn{2}{c}{1-shot}             & \multicolumn{2}{c}{2-shot} & \multicolumn{2}{c}{4-shot} & \multicolumn{2}{c}{5-shot} & \multicolumn{2}{c}{10-shot}                                    \\
      \cmidrule(lr){3-4}
      \cmidrule(lr){5-6}
      \cmidrule(lr){7-8}
      \cmidrule(lr){9-10}
      \cmidrule(lr){11-12}

                                                            & category   & mean                                   & std                        & mean                       & std                        & mean                        & std  & mean  & std & mean  & std \\
      \midrule
      \multirow[c]{15}{*}{\rotatebox[origin=c]{90}{\mvt/}}  & bottle     & 99.4                                   & 0.4                        & 99.3                       & 0.4                        & 99.4                        & 0.3  & 99.8  & 0.3 & 99.7  & 0.4 \\
                                                            & cable      & 71.7                                   & 6.1                        & 77.3                       & 3.6                        & 84.2                        & 8.4  & 82.9  & 3.3 & 87.0  & 3.6 \\
                                                            & capsule    & 78.6                                   & 12.9                       & 90.1                       & 4.8                        & 86.4                        & 11.8 & 94.3  & 4.6 & 92.8  & 4.5 \\
                                                            & carpet     & 97.0                                   & 0.5                        & 97.4                       & 0.6                        & 97.2                        & 0.2  & 97.2  & 0.4 & 97.5  & 0.7 \\
                                                            & grid       & 65.5                                   & 6.4                        & 74.1                       & 3.5                        & 86.0                        & 5.9  & 90.2  & 4.8 & 93.7  & 3.8 \\
                                                            & hazelnut   & 91.1                                   & 4.1                        & 95.7                       & 3.3                        & 98.7                        & 1.0  & 98.8  & 0.8 & 99.9  & 0.2 \\
                                                            & leather    & 100.0                                  & 0.0                        & 100.0                      & 0.0                        & 100.0                       & 0.0  & 100.0 & 0.0 & 100.0 & 0.0 \\
                                                            & metal-nut  & 79.0                                   & 2.4                        & 88.0                       & 5.6                        & 90.7                        & 5.1  & 97.0  & 0.8 & 99.1  & 0.6 \\
                                                            & pill       & 83.1                                   & 2.3                        & 85.5                       & 2.0                        & 87.6                        & 1.5  & 89.7  & 2.5 & 91.0  & 1.8 \\
                                                            & screw      & 49.2                                   & 2.8                        & 53.6                       & 4.5                        & 59.4                        & 3.9  & 62.1  & 3.7 & 70.5  & 4.9 \\
                                                            & tile       & 99.0                                   & 0.8                        & 98.8                       & 1.1                        & 98.8                        & 0.5  & 98.9  & 0.4 & 98.9  & 0.3 \\
                                                            & toothbrush & 93.1                                   & 2.6                        & 96.1                       & 2.7                        & 97.2                        & 3.4  & 99.4  & 0.8 & 99.9  & 0.2 \\
                                                            & transistor & 76.6                                   & 14.6                       & 81.4                       & 13.3                       & 95.1                        & 0.3  & 95.0  & 0.5 & 96.4  & 0.2 \\
                                                            & wood       & 99.5                                   & 0.1                        & 99.4                       & 0.2                        & 99.5                        & 0.2  & 99.4  & 0.1 & 99.5  & 0.1 \\
                                                            & zipper     & 96.2                                   & 1.6                        & 97.5                       & 2.2                        & 98.9                        & 0.4  & 98.9  & 0.8 & 99.2  & 0.2 \\
      \midrule
      \multirow[c]{12}{*}{\rotatebox[origin=c]{90}{\visa/}} & candle     & 93.4                                   & 1.4                        & 93.5                       & 1.0                        & 93.6                        & 1.3  & 93.6  & 0.6 & 95.0  & 1.0 \\
                                                            & capsules   & 64.2                                   & 6.8                        & 66.3                       & 4.1                        & 70.1                        & 3.6  & 70.3  & 1.8 & 76.2  & 1.7 \\
                                                            & cashew     & 93.7                                   & 1.4                        & 95.3                       & 1.2                        & 96.7                        & 0.6  & 96.8  & 0.7 & 96.6  & 0.7 \\
                                                            & chewinggum & 97.8                                   & 0.5                        & 97.7                       & 0.2                        & 98.4                        & 0.5  & 97.9  & 0.3 & 97.9  & 0.5 \\
                                                            & fryum      & 84.1                                   & 1.2                        & 80.1                       & 4.3                        & 89.1                        & 2.4  & 92.0  & 1.6 & 93.2  & 1.1 \\
                                                            & macaroni1  & 75.6                                   & 3.2                        & 80.7                       & 4.6                        & 86.4                        & 2.9  & 87.3  & 5.7 & 92.3  & 2.2 \\
                                                            & macaroni2  & 58.9                                   & 8.8                        & 59.4                       & 4.8                        & 61.2                        & 3.1  & 64.3  & 7.8 & 66.7  & 2.5 \\
                                                            & pcb1       & 80.3                                   & 19.5                       & 90.5                       & 3.7                        & 93.1                        & 3.8  & 94.7  & 2.4 & 96.8  & 0.9 \\
                                                            & pcb2       & 88.5                                   & 1.6                        & 91.3                       & 0.6                        & 92.9                        & 0.8  & 92.3  & 0.9 & 93.4  & 0.8 \\
                                                            & pcb3       & 83.0                                   & 2.3                        & 85.9                       & 2.0                        & 88.3                        & 1.4  & 90.3  & 2.5 & 91.8  & 1.4 \\
                                                            & pcb4       & 86.5                                   & 11.0                       & 96.2                       & 2.1                        & 97.0                        & 2.3  & 98.5  & 1.1 & 98.7  & 0.4 \\
                                                            & pipe-fryum & 97.6                                   & 1.3                        & 98.8                       & 0.3                        & 99.0                        & 0.1  & 99.2  & 0.2 & 99.3  & 0.1 \\
      \bottomrule
   \end{tabular}
\end{table*}

\end{document}